\PassOptionsToPackage{table}{xcolor}
\documentclass[]{gensi}

\usepackage{amsmath,amsfonts,bm}

\def\eqref#1{equation~\ref{#1}}

\def\1{\bm{1}}

\DeclareMathAlphabet{\mathsfit}{\encodingdefault}{\sfdefault}{m}{sl}
\SetMathAlphabet{\mathsfit}{bold}{\encodingdefault}{\sfdefault}{bx}{n}

\usepackage{float}
\usepackage{xurl}
\usepackage{tabularx}
\usepackage{booktabs}
\usepackage[toc,page,header]{appendix}
\usepackage{algorithm}
\usepackage{algorithmic}
\usepackage{mdframed}
\usepackage{siunitx}
\usepackage{fix-cm}
\usepackage{caption}

\newcommand{\method}{Direct-OPD}

\newcommand{\gain}[1]{\textcolor{green!55!black}{\scriptsize #1}}
\newcommand{\takeaway}[1]{%
  \begin{center}
  \begingroup
  \setlength{\fboxsep}{7pt}%
  \fcolorbox{black!18}{blue!4}{%
    \parbox{0.88\linewidth}{\textbf{Takeaway.} #1}%
  }%
  \endgroup
  \end{center}
}

\title{Weak-to-Strong Generalization via Direct On-Policy Distillation}
\author[1, 2, 4]{Shiyuan Feng\textsuperscript{*}}
\author[1, 2, 3]{Huan-ang Gao\textsuperscript{*, $\ddagger$}}
\author[1, 2, 3]{Haohan Chi\textsuperscript{*}}
\author[1, 2]{Hanlin Wu}
\author[1, 2]{Zhilong Zhang}
\author[3]{Zheng Jiang}
\author[3]{Bingxiang He}
\author[1, 2]{Wei-Ying Ma}
\author[1, 2]{Ya-Qin Zhang}
\author[1, 2]{Hao Zhou\textsuperscript{$\dagger$}}

\affiliation[1]{SIA-Lab of Tsinghua AIR and ByteDance Seed}
\affiliation[2]{Institute for AI Industry Research (AIR), Tsinghua University}
\affiliation[3]{Department of Computer Science and Technology, Tsinghua University}
\affiliation[4]{Peking University}

\date{\today}
\correspondence{\url{zhouhao@air.tsinghua.edu.cn}}
\projectpage{\url{https://bytedtsinghua-sia.github.io/Direct-OPD/}}

\abstract{%
Reinforcement learning with verifiable rewards (RLVR) is a powerful recipe for
improving language-model reasoning, but it is expensive to repeat on every new
strong model because the target model must generate many rollouts during
training. As models scale, post-training itself becomes a bottleneck. We study a
weak-to-strong alternative: run RL on a smaller model where rollouts are cheaper,
then reuse what that RL run learned to improve a stronger target model. Directly
distilling the post-RL weak teacher is not enough, because the teacher's final
policy mixes useful RL gains with the limitations of the smaller model. We
propose \textbf{Direct On-Policy Distillation} (\textbf{\method{}}), which
transfers the teacher's RL-induced \textbf{policy shift} instead. \method{}
compares the post-RL teacher with its own pre-RL reference and treats their
log-ratio as a dense implicit reward for the student. In plain terms, the
checkpoint pair tells us which actions RL made the weak model more or less likely
to take, and \method{} applies that signal on the stronger student's own
on-policy states. This directly reuses the weak model’s RL supervision signal without running sparse-reward RL on the target model. Empirically, \method{} consistently leverages weaker teachers to
improve stronger target models; notably, it boosts Qwen3-1.7B \uline{from
48.3\% to 58.3\% on AIME 2024 in just 4 hours on 8 A100 GPUs}. 
It outperforms
step-matched direct RL and enables the sequential composition of multiple policy
shifts. Our results show that RL outcomes can be reused across model scales as
implicit reward signals, not merely as final models to imitate.
}

\begin{document}

\maketitle

\begingroup
\renewcommand{\thefootnote}{}
\footnotetext{
\textsuperscript{*}Equal contribution. \quad
\textsuperscript{$\ddagger$}Project Lead. \quad
\textsuperscript{$\dagger$}Corresponding author.
}
\endgroup

\section{Introduction}

\begin{figure}[t]
\centering
\includegraphics[width=\linewidth]{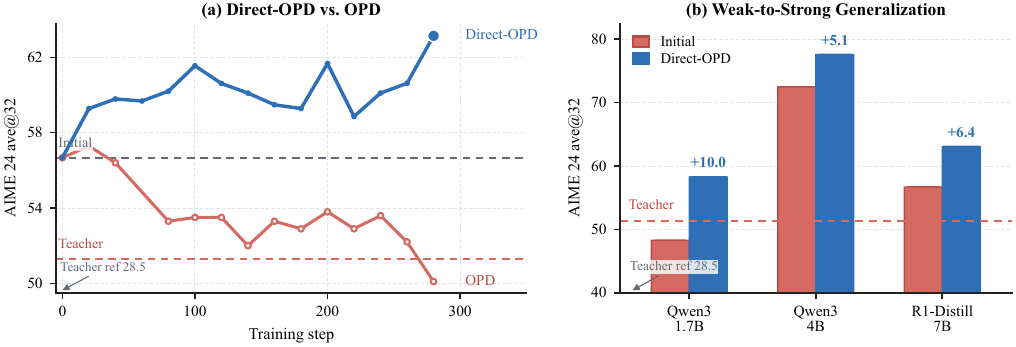}
\caption{\textbf{\method{} transfers the effect of small-model RL rather than
imitating the small model.} (a) Starting from R1-Distill-7B, vanilla OPD toward
the post-RL JustRL-1.5B teacher degrades performance, whereas \method{} transfers
the JustRL-1.5B $-$ R1-Distill-1.5B policy shift and improves the student.
(b) The same policy shift improves Qwen3-1.7B, Qwen3-4B, and R1-Distill-7B on
AIME 2024, including students whose initial accuracy already exceeds the
post-RL teacher.}
\label{fig:intro-opd-transfer}
\end{figure}

Reinforcement learning with verifiable rewards (RLVR) has become a dominant
post-training recipe for eliciting strong reasoning in large language
models~\citep{deepseekr1,justrl}. Its cost, however, is tied to the
model being trained: every update requires the current policy to generate
rollouts, receive verifiable outcome scores, and update on those trajectories.
Larger target models make each rollout slower and each RL iteration more
expensive. Thus, as reasoning models continue to scale, re-running RLVR from scratch
for every new strong model risks making post-training itself a bottleneck.

In this paper, we propose \textbf{Direct On-Policy Distillation}
(\textbf{\method{}}), a weak-to-strong post-training paradigm. The setting is
simple: run RL where it is cheap, on a weak model, and use what that RL run
learned to train a stronger target model.
Crucially, the weak model is not assumed to be a better reasoner than the student; it is merely an inexpensive vehicle which ships the behavior the RL signal has already reshaped. \method{} transfers the improvement induced by weak-model RL, not the weak model itself.

The central question is what exactly to transfer from the weak RL run. One candidate is the post-RL teacher's final policy: on-policy distillation (OPD) lets the student sample its own states and trains it to match the teacher there. Yet this is the wrong object for weak-to-strong transfer, because the final policy entangles the useful change induced by RL with the weak model's intrinsic capacity limits. When the student already surpasses the teacher, imitating that policy can overwrite stronger behavior. Figure~\ref{fig:intro-opd-transfer}(a) illustrates this failure: R1-Distill-7B~\citep{deepseekr1} begins at $56.7$, already above the post-RL JustRL-1.5B teacher~\citep{justrl} at $51.3$, yet vanilla OPD drags it down to roughly $50$. The weak RL run carries useful supervision, but its final policy is a poor carrier of that supervision.

\method{} instead isolates the supervision by contrasting the weak model with itself, before and after RL. Let $\pi_{T_{\mathrm{ref}}}$ denote the weak model before RL and $\pi_T$ its post-RL counterpart. We define the teacher policy shift
\begin{equation}
    \Delta_T(y \mid x)
    =
    \log \pi_T(y \mid x)
    -
    \log \pi_{T_{\mathrm{ref}}}(y \mid x),
\end{equation}
which is positive on responses that RL made more likely and negative on those it suppressed. The subtraction discards what the weak model already preferred before RL and retains only what RL changed. More importantly, under the KL-regularized RL objective, this policy shift is mathematically equivalent to the reward that trained the weak model: optimizing against $\Delta_T$ recovers the same objective form as optimizing against that reward, as proved in Section~\ref{sec:reward}. A pair of weak-model checkpoints therefore stores the RL supervision signal directly in policy space. \method{} applies this signal on the student's own on-policy states, anchored to the student initialization by a KL term, without training an explicit reward model or running sparse-reward RL on the target.

This mechanism converts weak-model RL into a transferable post-training signal. Figure~\ref{fig:intro-opd-transfer}(b) shows that the policy shift from R1-Distill-1.5B to JustRL-1.5B improves Qwen3-1.7B, Qwen3-4B~\citep{qwen3technicalreport}, and R1-Distill-7B alike, including students that already start above the post-RL teacher. The transfer is also inexpensive: with 8 A100 GPUs for about 4 hours, \method{} raises Qwen3-1.7B from $48.3$ to $58.3$ on AIME 2024, comparable to what Polaris attains by running RL directly on Qwen3-1.7B with 32 A100 GPUs for at least a week~\citep{polaris}.

\paragraph{Contributions.}
\begin{itemize}
    \item \textbf{A weak-to-strong post-training paradigm.} We recast small-model RL as a cheap generator of an implicit reward for stronger models: rather than copying a weak teacher's final policy, we evaluate its RL-induced policy shift on the student's own on-policy states.
    \item \textbf{Consistent gains across teacher pairs and students.} With
    two teacher pairs (R1-Distill-1.5B $\rightarrow$ JustRL-1.5B and
    Nemotron-1.5B $\rightarrow$ QuestA-Nemotron-1.5B) and three students from
    two families (R1-Distill-7B, Qwen3-1.7B, and Qwen3-4B), \method{} improves every
    student---including R1-Distill-7B and Qwen3-4B, which already start above the
    post-RL teacher---and lifts Qwen3-1.7B from $48.3$ to $58.3$ on AIME 2024 in
    about 4 hours on 8 A100 GPUs. 
    \item \textbf{Cheaper than direct large-scale RL.} At matched RL steps,
    running RL on a 1.5B model and transferring with \method{} outperforms
    running RL directly on R1-Distill-7B in both accuracy and compute.
    \item \textbf{Analysis of when the signal is reliable.} Unlike imitating the
    teacher's final policy, \method{} does not require high teacher--student
    top-$k$ overlap and transfers
    across thinking patterns; we identify the response-length and KL conditions
    under which the implicit reward stays aligned with validation accuracy.
\end{itemize}

\section{Direct On-Policy Distillation}
\label{sec:method}



\subsection{Preliminaries: On-Policy Distillation}
\label{sec:preliminaries}

We consider autoregressive language-model policies. Let $x\sim\mathcal{D}$ be a
prompt and $y=(y_1,\ldots,y_T)$ a response, so that a policy $\pi$ factorizes as
$\log\pi(y\mid x)=\sum_{t=1}^{T}\log\pi(y_t\mid s_t)$ over prefixes
$s_t=(x,y_{<t})$. We distinguish four policies: the policy being trained
$\pi_\theta$ and its initial checkpoint $\pi_S$; and a teacher pair consisting of
a pre-RL reference $\pi_{T_{\mathrm{ref}}}$ and the corresponding post-RL teacher
$\pi_T$. Our weak-to-strong setting is the regime in which $\pi_T$ may be smaller or weaker than the target student, yet the transition from $\pi_{T_{\mathrm{ref}}}$ to $\pi_T$ still encodes information worth transferring.

On-policy distillation (OPD) supervises the student on trajectories it samples
itself. Given a prompt $x$, the student draws $\hat y\sim\pi_\theta(\cdot\mid x)$,
and at each visited prefix $s_t=(x,\hat y_{<t})$ we read off the student and
teacher next-token distributions $p_t(v)=\pi_\theta(v\mid s_t)$ and
$q_t(v)=\pi_T(v\mid s_t)$. Standard OPD minimizes the sequence-level KL to the
teacher on these rollouts,
\begin{equation}
\mathcal{L}_{\mathrm{OPD}}(\theta)
=\mathbb{E}_{x\sim\mathcal{D}}\big[\,
D_{\mathrm{KL}}\big(\pi_\theta(\cdot\mid x)\,\big\|\,\pi_T(\cdot\mid x)\big)\,\big],
\end{equation}
which factorizes exactly into dense, per-token supervision on the states the
student actually visits,
\begin{equation}
\mathcal{L}_{\mathrm{OPD}}(\theta)
=\mathbb{E}_{x\sim\mathcal{D},\,\hat y\sim\pi_\theta}\Big[\,
\textstyle\sum_{t=1}^{T}D_{\mathrm{KL}}(p_t\,\|\,q_t)\,\Big].
\end{equation}
In practice the teacher is queried only on the student's top-$k$ support
$S_t=\operatorname{TopK}_v p_t$, giving the local estimator
\begin{equation}
\mathcal{L}^{\text{top-}k}_{\mathrm{OPD}}(\theta)
=\mathbb{E}_{x\sim\mathcal{D},\,\hat y\sim\pi_\theta}\Big[\,
\textstyle\sum_{t=1}^{T}
D_{\mathrm{KL}}\big(\bar p_t^{S_t}\,\big\|\,\bar q_t^{S_t}\big)\,\Big],
\end{equation}
where $\bar p_t^{S_t}$ and $\bar q_t^{S_t}$ renormalize $p_t$ and $q_t$ on
$S_t$~\citep{li2026rethinkingopd}. This top-$k$ KL estimator produces a dense
per-token training signal involving the teacher--student log-ratio
$\log q_t(v)-\log p_t(v)$. \method{} inherits this on-policy, top-$k$ interface unchanged, and departs from OPD only in \textit{what signal is read from the teacher} 
(Section~\ref{sec:reward}).

\subsection{Policy Shifts as Implicit Rewards}
\label{sec:reward}

\textbf{Which signal to transfer.} Rather than imitating the teacher's absolute output distribution, \method{} transfers its RL-induced \textbf{policy shift}: the sequence-level log-ratio between the post-RL teacher $\pi_T$ and its pre-RL reference $\pi_{T_{\mathrm{ref}}}$,
\begin{equation}
    \Delta_T(y\mid x) = \log \pi_T(y\mid x) - \log \pi_{T_{\mathrm{ref}}}(y\mid x),
    \label{eq:dopd-delta}
\end{equation}
which isolates the teacher's RL-induced \textit{direction} rather than its
absolute final distribution. This shift is not an arbitrary heuristic: under
the \textbf{policy-as-reward} view of KL-regularized RL, it can be interpreted
as the teacher's implicit reward.

For a reward $r$, reference policy $\pi_{\mathrm{ref}}$, and penalty $\beta>0$,
the KL-regularized objective
$\max_\pi\mathbb{E}_{y\sim\pi}\!\big[r(x,y)-\beta\log\tfrac{\pi(y\mid x)}{\pi_{\mathrm{ref}}(y\mid x)}\big]$
admits the closed-form optimum $\pi^*\propto\pi_{\mathrm{ref}}\exp(r/\beta)$, hence
\begin{equation}
\log\frac{\pi^*(y\mid x)}{\pi_{\mathrm{ref}}(y\mid x)}
=\tfrac{1}{\beta}\,r(x,y)-\log Z(x).
\end{equation}
Since $\log Z(x)$ is constant across responses to the same prompt, the
policy--reference log-ratio recovers the reward up to a positive scale and a
per-prompt constant. This is the same identity that underlies Direct Preference
Optimization, which fits a policy from preferences without a separately trained
reward model~\citep{rafailov2023direct}. Here we use the identity in the
opposite direction: we are already \textit{given} a post-RL teacher $\pi_T$ and
its pre-RL reference $\pi_{T_{\mathrm{ref}}}$, and we read the reward-like signal
back out of their policy ratio. Treating $\pi_T$ as the KL-regularized optimum of some latent reward $r_T$ anchored at $\pi_{T_{\mathrm{ref}}}$,
\begin{equation}
    \Delta_T(y\mid x) = \tfrac{1}{\beta}\, r_T(x,y) - \log Z_T(x),
    \label{eq:dopd-implicit-reward}
\end{equation}
so the shift $\Delta_T$ can be interpreted as the teacher's \textbf{implicit
reward}, up to a positive scale and a per-prompt constant. \method{} transfers
this reward-like signal to the student.

\subsection{The Direct-OPD Objective}
\label{sec:dopd}

\textbf{Idealized sequence-level objective.}
We first write \method{} as a sequence-level objective. With the student
initialized from $\pi_S$, \method{} optimizes the student against the
teacher-shift signal $\Delta_T$ while regularizing the student toward its own
initialization,
\begin{equation}
    J_{\text{\method}}(\theta) = \mathbb{E}_{x\sim\mathcal{D}} \left[
    \mathbb{E}_{y\sim\pi_\theta(\cdot\mid x)} \big[ \Delta_T(y\mid x) \big]
    - \alpha\, D_{\mathrm{KL}} \big( \pi_\theta(\cdot\mid x) \,\|\, \pi_S(\cdot\mid x) \big) \right],
    \label{eq:dopd-obj}
\end{equation}
where $\alpha>0$ controls the KL penalty. In this idealized form, the objective
is itself a KL-regularized RL objective on the student: its optimum is
\begin{equation}
    \pi^*(y\mid x) \propto \pi_S(y\mid x)\, \exp\!\big(\tfrac{1}{\alpha}\Delta_T(y\mid x)\big)
    = \pi_S(y\mid x) \left( \frac{\pi_T(y\mid x)}{\pi_{T_{\mathrm{ref}}}(y\mid x)} \right)^{1/\alpha},
    \label{eq:dopd-opt}
\end{equation}
and substituting the implicit-reward form of Eq.~\ref{eq:dopd-implicit-reward}
gives $\pi^*\propto\pi_S\exp(r_T/\alpha\beta)$. The student is therefore
optimized as if it were running KL-regularized RL with the small teacher's
implicit reward, while using its own initialization $\pi_S$ as the reference.
Crucially, it obtains this reward-like signal entirely from the checkpoint pair, without querying a verifiable reward or running sparse-reward RL on the target.

\textbf{From sequence to token level.}
Because both teachers factorize autoregressively over the same prefixes, the
sequence shift decomposes exactly into a sum of token-level shifts,
\begin{equation}
    \Delta_T(y\mid x) = \sum_{t} r_t(y_t\mid s_t),
    \qquad
    r_t(v) = \log \pi_T(v\mid s_t) - \log \pi_{T_{\mathrm{ref}}}(v\mid s_t),
    \label{eq:dopd-token-reward}
\end{equation}
so that $r_t(v)$ can be viewed as a \emph{dense immediate per-token reward}:
$r_t(v)>0$ where the teacher's RL encouraged token $v$, and $r_t(v)<0$ where it
suppressed it. In practice, we use the corresponding zero-discount token-level
surrogate for Eq.~\ref{eq:dopd-obj}, crediting each candidate token by its
immediate shift $r_t(v)$ rather than by a future return over later shifts. This
reuses the on-policy, top-$k$ interface of Section~\ref{sec:preliminaries}.

\textbf{Top-$k$ action-space restriction.}
The implementation uses a local top-$k$ approximation to this zero-discount
token-level surrogate. To keep the signal on actions the student actually
considers, at each visited prefix we restrict to its top-$k$ support
$S_t=\operatorname{TopK}_v \pi_\theta(v\mid s_t)$ and renormalize the student
probabilities on it,
\begin{equation}
    \bar p_t(v) =
    \frac{\pi_\theta(v\mid s_t)}
    {\sum_{u\in S_t}\pi_\theta(u\mid s_t)},
    \qquad v\in S_t,
\end{equation}
which is the renormalized student distribution $\bar p_t^{S_t}$ of
Section~\ref{sec:preliminaries}.

\textbf{Analytical top-$k$ policy gradient.}
The immediate per-token reward $r_t(v)$ depends only on the teacher pair, so we
use it directly as the local advantage. A single-sample, token-level policy
gradient for the zero-discount surrogate would weight the log-likelihood of the
\emph{sampled} token by its immediate reward,
\begin{equation}
    \nabla_\theta J_{\mathrm{MC}} =
    \mathbb{E}_{x, y} \Big[
    \textstyle\sum_t r_t(y_t)\,
    \nabla_\theta \log \pi_\theta(y_t \mid s_t)
    \Big],
\end{equation}
but estimating each step from one token is high-variance. Since we already have
the full top-$k$ rewards and probabilities at every visited prefix, we replace
the single-token estimate at each step by its expectation over the restricted
distribution $\bar p_t$, i.e. we Rao--Blackwellize the per-step action while
still sampling the trajectory $y\sim\pi_\theta$:
\begin{equation}
    \nabla_\theta J_{\mathrm{analytical}} =
    \mathbb{E}_{x,\, y\sim\pi_\theta} \Big[
    \sum_t \sum_{v \in S_t}
    \underbrace{\textcolor{blue}{\bar p_t(v)}}_{\text{weight}} \,
    \underbrace{\textcolor{red}{r_t(v)}}_{\text{reward}} \,
    \nabla_\theta
    \underbrace{\textcolor{magenta}{\log \pi_\theta(v \mid s_t)}}_{\text{log-likelihood}}
    \Big].
\end{equation}
This gives a Rao--Blackwellized per-step estimator under the restricted top-$k$
action distribution; it removes the token-sampling variance of
$\nabla_\theta J_{\mathrm{MC}}$ while leaving the sampled trajectory distribution
untouched.

\textbf{Stop-gradient coefficient.}
For this surrogate to match the policy-gradient form, the weight $\bar p_t(v)$
must enter only as a scalar. It depends on $\theta$ through the student softmax,
so differentiating it would inject extra Jacobian terms from the top-$k$
distribution. We therefore detach the weighted reward into a static coefficient,
\begin{equation}
    A_t^{\mathrm{w}}(v) =
    \operatorname{stop\_gradient}\!\big(
    \textcolor{blue}{\bar p_t(v)} \cdot
    \textcolor{red}{r_t(v)}
    \big),
\end{equation}
giving the final local top-$k$ surrogate for the zero-discount objective,
\begin{equation}
    \nabla_\theta J_{\text{\method}} \approx
    \mathbb{E}_{x \sim \mathcal{D},\, y \sim \pi_\theta} \Big[
    \sum_t \sum_{v \in S_t}
    A_t^{\mathrm{w}}(v)\,
    \nabla_\theta
    \textcolor{magenta}{\log \pi_\theta(v \mid s_t)}
    \Big]
    - \alpha\, \nabla_\theta
    D_{\mathrm{KL}}\big(
    \pi_\theta(\cdot\mid x) \,\|\, \pi_S(\cdot\mid x)
    \big).
\end{equation}
In practice we do not differentiate the KL analytically; we use the standard
KL-penalty implementation in \texttt{verl} to anchor $\pi_\theta$ to $\pi_S$.

\subsection{Adaptive KL Control}
\label{sec:method-adaptive-kl}

The reward \method{} transfers carries a scale it does not get to choose. By
Eq.~\ref{eq:dopd-implicit-reward}, $\Delta_T=\tfrac{1}{\beta}r_T-\log Z_T$: its magnitude is the teacher's reward divided by the teacher's KL budget $\beta$, both fixed when the teacher was trained and encoded in how far its RL pushed $\pi_T$ from $\pi_{T_{\mathrm{ref}}}$, yet neither is recoverable from the checkpoint pair.
The per-token shift $r_t(v)=\log\pi_T(v\mid s_t)-\log\pi_{T_{\mathrm{ref}}}
(v\mid s_t)$ is likewise a function of the teacher pair alone; the student enters
only through \emph{where} the signal is read off---the prefixes its rollouts
visit and the top-$k$ actions retained at each one.

This scale mismatch makes a \emph{fixed} $\alpha$ unlikely to transfer robustly
across settings. From the optimum in Eq.~\ref{eq:dopd-opt},
$\pi^*\propto\pi_S\exp(\tfrac{1}{\alpha}\Delta_T)$, $\alpha$ is the temperature
of an exponential tilt of $\pi_S$, and is meaningful only relative to the scale
of $\Delta_T$. But $\Delta_T$ lives in the teacher's reward units
($\tfrac1\beta r_T$) while $\alpha$ is a coefficient on the \emph{student's}
KL---two scales with no a priori conversion, the teacher's being unobservable.
A single $\alpha$ therefore cannot be calibrated in advance across teacher--student pairs, and we instead adopt a lightweight controller that adapts $\alpha$ to the running scale of the dense signal.

Let $\alpha_m$ be the KL coefficient at training iteration $m$, and let
$\bar r_m$ be the batch-level mean of the student-weighted shift
$\bar p_t(v)\,r_t(v)$ over the visited prefixes and their top-$k$ candidates---the
same dense reward the gradient optimizes. Before the actor update we set
\begin{equation}
    \alpha_{m+1}
    =
    \operatorname{clip}
    \left(
    \alpha_m
    \left(1 + \epsilon\,\operatorname{sgn}(\bar r_m)\right),
    \alpha_{\min},
    \alpha_{\max}
    \right),
    \label{eq:adaptive-kl}
\end{equation}
with $\operatorname{sgn}(0)=0$; by default $\epsilon=0.01$ and
$[\alpha_{\min},\alpha_{\max}]=[0.5,2.5]$. Here $\operatorname{sgn}(\bar r_m)$
reports whether, on the prefixes the student currently visits, the teacher's RL
on average raised or lowered the retained candidate tokens---a batch-level sign
that the student-weighted shift carries directly.

The controller is intended to keep the dense signal balanced. When the visited
prefixes carry positive teacher shift on average, it \emph{raises} $\alpha$ to
discourage over-amplifying that local signal; when the average shift is
negative, it \emph{lowers} $\alpha$, weakening the anchor so the dense gradient
can move probability away from tokens whose likelihood the teacher's RL reduced.
This differs from the standard adaptive controller for an in-reward KL penalty,
which steers toward a target KL value: here the update is driven by the
\emph{sign of the \method{} dense reward} and acts on the explicit actor KL loss
coefficient. We analyze KL-coefficient sensitivity in
Section~\ref{sec:kl-coefficient}.

\section{Experiments}

In this section we answer the following research questions:
\begin{itemize}
    \item \textbf{RQ1.} Can a small teacher's RL-induced policy shift improve
    students that already match or exceed the teacher's own ability, and does
    this hold across different teacher pairs and student families?
    (Section~\ref{sec:small-teacher-stronger})
    \item \textbf{RQ2.} Under a fixed RL-step budget, does running RL on a small model
    and transferring its policy shift with \method{} outperform running RL
    directly on the large target---in both accuracy and compute?
    (Section~\ref{sec:weak-to-strong-beats-rl})
    \item \textbf{RQ3.} Can several independently learned policy shifts be
    composed sequentially to accumulate their gains in a single student?
    (Section~\ref{sec:sequential-composition})
\end{itemize}

\subsection{A small RL teacher improves much stronger students}
\label{sec:small-teacher-stronger}

Our first experiment asks whether an RL improvement learned by a small teacher can be used to improve students that are stronger than the post-RL teacher. We use R1-Distill-1.5B as the teacher reference and JustRL-1.5B as the post-RL teacher, then distill the policy shift from this pair into three target students: R1-Distill-7B, Qwen3-1.7B, and Qwen3-4B. In this setting, R1-Distill-7B and Qwen3-4B already outperform the post-RL JustRL teacher before Direct-OPD, making it a direct weak-to-strong transfer test.

We further test a second teacher pair, Nemotron-1.5B → QuestA-Nemotron-1.5B, which comes from a different training pipeline and data source. This setting is not intended as a strict weaker-teacher-to-stronger-student comparison, since the post-RL QuestA teacher is already strong on AIME. Instead, it tests whether Direct-OPD can transfer a policy-shift signal beyond a single teacher family, training recipe, and data source. We transfer this policy shift into R1-Distill-7B and Qwen3-1.7B.

\begin{figure}[H]
\centering
\begin{minipage}[t]{0.70\linewidth}
\vspace{0pt}
\centering
\includegraphics[width=\linewidth]{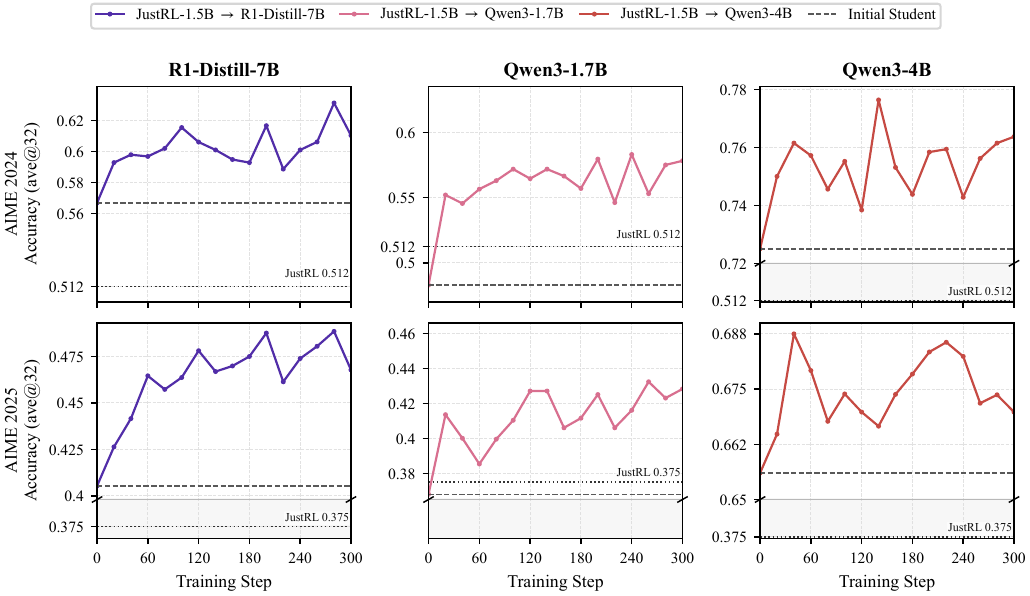}
\end{minipage}\hfill
\begin{minipage}[t]{0.208\linewidth}
\vspace{1.8em}
\centering
\includegraphics[width=\linewidth]{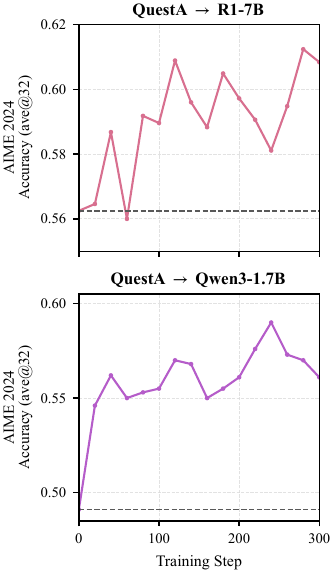}
\end{minipage}
\caption{\method{} transfers RL-induced policy shifts across teacher pairs and
student families. \textbf{Left:} R1-Distill-1.5B $\rightarrow$ JustRL-1.5B
transfer into R1-Distill-7B, Qwen3-1.7B, and Qwen3-4B, evaluated on AIME 2024
and AIME 2025. \textbf{Right:} Nemotron-1.5B $\rightarrow$
QuestA-Nemotron-1.5B transfer into R1-Distill-7B and Qwen3-1.7B on AIME 2024.
The two teacher pairs come from different training data and pipelines, showing
that \method{} is not specific to a single post-RL teacher.}
\label{fig:transfer-generalization}
\end{figure}

Figure~\ref{fig:transfer-generalization} shows that the transferred policy shift improves students across these settings, including R1-Distill-7B and Qwen3-4B students whose starting accuracies already exceed the post-RL JustRL teacher. The QuestA results serve as an additional robustness check: they show that the effect is not specific to the JustRL teacher pair, and can also appear with a different teacher training pipeline, data source, and post-RL checkpoint. Taken together, these results suggest that Direct-OPD transfers a reusable policy-shift signal rather than simply imitating the post-RL teacher’s final policy.

This result supports the interpretation that \method{} uses an RL-induced
direction of improvement rather than the teacher's full policy.

\begin{table}[H]
\centering
\small
\setlength{\tabcolsep}{4.2pt}
\renewcommand{\arraystretch}{1.07}
\begin{minipage}[t]{0.49\linewidth}
\centering
\textbf{(a) JustRL policy-shift transfer}\\[-1pt]
\begin{tabular}{lcc}
\toprule
Model & AIME24 & AIME25 \\
\midrule
Teacher ref & 28.5 & 24.0 \\
Teacher RL~\citep{justrl} & 51.3 & 37.5 \\
\midrule
Qwen3-1.7B & 48.3 & 36.8 \\
\rowcolor{cyan!12}
+ \method{} & \textbf{58.3} \gain{+10.0} & \textbf{43.2} \gain{+6.4} \\
Qwen3-4B & 72.5 & 65.6 \\
\rowcolor{cyan!12}
+ \method{} & \textbf{77.6} \gain{+5.1} & \textbf{68.8} \gain{+3.2} \\
R1-Distill-7B & 56.7 & 40.5 \\
\rowcolor{cyan!12}
+ \method{} & \textbf{63.1} \gain{+6.4} & \textbf{48.8} \gain{+8.3} \\
\bottomrule
\end{tabular}
\end{minipage}\hfill
\begin{minipage}[t]{0.47\linewidth}
\centering
\textbf{(b) QuestA policy-shift transfer}\\[-1pt]
\begin{tabular}{lcc}
\toprule
Model & AIME24 & AIME25 \\
\midrule
Teacher ref & 61.77 & 49.50 \\
Teacher RL~\citep{questa} & 72.50 & 62.29 \\
\midrule
Qwen3-1.7B & 48.3 & 36.8 \\
\rowcolor{cyan!12}
+ \method{} & \textbf{59.0} \gain{+10.7} & \textbf{43.1} \gain{+6.3} \\
R1-Distill-7B & 56.3 & 39.5 \\
\rowcolor{cyan!12}
+ \method{} & \textbf{61.2} \gain{+4.9} & \textbf{44.0} \gain{+4.5} \\
\bottomrule
\end{tabular}
\end{minipage}
\caption{Score summaries for the different students and teacher pairs}
\label{tab:transfer-score-summary}
\end{table}

\subsection{Weak-to-strong generalization beats RL}
\label{sec:weak-to-strong-beats-rl}

We next ask whether \method{} is only a way to reuse an already trained small
teacher, or whether it can be a better training path than running RL directly
on the larger target model. We compare two matched-step routes. The direct-RL
route trains R1-Distill-7B with RL. The weak-to-strong route first trains the
smaller R1-Distill-1.5B model with RL, then transfers the resulting
RL/reference policy shift into R1-Distill-7B using \method{}.

We test that for the same number of RL steps, a smaller model is a more efficient place to
find a useful policy-improvement direction, and \method{} can then transfer that
direction to the larger student. We trained R1-Distill-1.5B on DAPO dataset, and selected R1-Distill-1.5B RL checkpoints at
steps 300, 600, 900, 1200, 1500. For each selected checkpoint, we construct a
teacher pair using the base R1-Distill-1.5B model as
$\pi_{T_{\mathrm{ref}}}$ and the selected RL checkpoint as $\pi_T$. We then
train R1-Distill-7B with \method{} and compare the
resulting validation score against direct R1-Distill-7B RL at the matched
small-teacher RL step. In the figure, T300 denotes the route that first trains
R1-Distill-1.5B with RL for 300 steps and then applies \method{} to
R1-Distill-7B; T600--T1500 are defined analogously.

We find that, for the same number of RL steps, running RL on the small model and
then transferring the learned policy shift to the large model gives better
R1-Distill-7B performance than running RL directly on R1-Distill-7B.

This matched-step advantage also translates into a compute advantage. In our
setup, a 1500-step RL run on R1-Distill-1.5B takes about 160 hours on
32 A100 GPUs, while RL on R1-Distill-7B takes
about 320 hours. After small-model RL, the \method{} transfer stage adds only about 4
hours on 8 A100 GPUs, which is negligible compared with the RL cost.
This low transfer cost comes from two factors: \method{} requires only a short
training run, as discussed in Section~\ref{sec:training-length}, and the teacher models used for
scoring are small enough that scoring is not the speed bottleneck.

\begin{figure}[H]
\centering
\includegraphics[width=\linewidth]{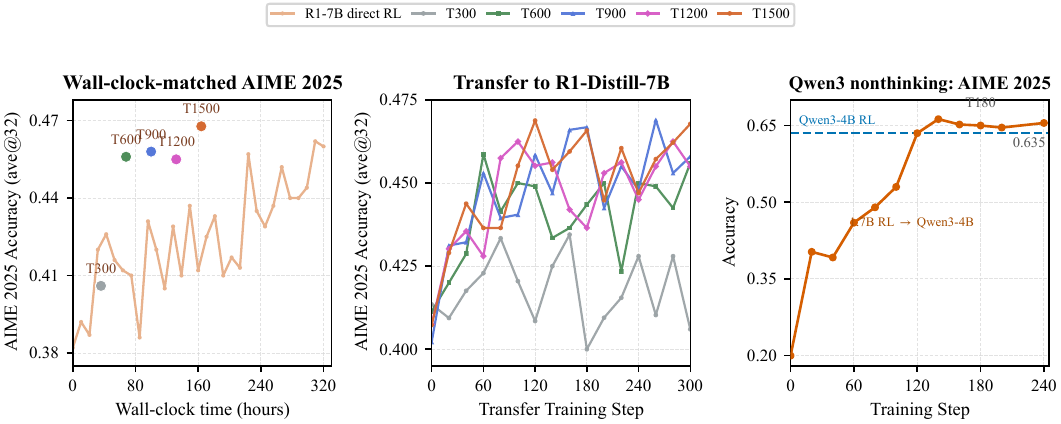}
\caption{\textbf{Running RL on a small model and transferring its policy shift
with \method{} beats running RL directly on the large target under a shorter
wall-clock training path.}
We compare two routes to improving R1-Distill-7B: \emph{direct RL} on
R1-Distill-7B, versus a \emph{weak-to-strong} route that runs RL on the smaller
R1-Distill-1.5B and transfers the resulting policy shift into R1-Distill-7B with
\method{}. T$N$ denotes the transfer that uses the R1-Distill-1.5B RL checkpoint
at step $N$ as the post-RL teacher $\pi_T$, with the base model as
$\pi_{T_{\mathrm{ref}}}$. \textbf{Left:} AIME 2025 accuracy against total
wall-clock time; the wiggly curve is direct R1-Distill-7B RL, and each T$N$
point sums the elapsed time for small-model RL and the short \method{} transfer.
Later transfers (T600--T1500) sit above the direct-RL curve at comparable
elapsed time. \textbf{Middle:} \method{} transfer trajectories into
R1-Distill-7B from the five small-teacher checkpoints; the early T300 carries a
weaker shift than T900--T1500. \textbf{Right:} the same recipe with Qwen3
nonthinking models---transferring a Qwen3-1.7B RL shift into Qwen3-4B reaches the
$0.635$ accuracy of direct Qwen3-4B RL (dashed) on AIME 2025.}
\label{fig:r1-7b-small-teacher-transfer}
\end{figure}

Figure~\ref{fig:r1-7b-small-teacher-transfer} shows that the transferred
signal depends on which small-teacher RL checkpoint is used. This dependence is
expected: different points in the small-model RL trajectory encode different
policy shifts, and not every shift is equally useful on the larger student's
state distribution.

This result changes the role of the small model. The small teacher is not used
because its endpoint policy is stronger than R1-Distill-7B. Instead, the small
model provides a cheaper environment in which RL discovers a direction of
improvement. \method{} then transfers that direction to the larger model by
evaluating the teacher/reference log-ratio on the larger student's own
on-policy states. Direct RL on the large model must discover the same kind of
direction through its own rollouts; weak-to-strong transfer reuses the
direction already found by the small model.

We observe the same pattern with Qwen3 nonthinking models. After a 100-step RL
run on the 1.7B model, transferring the resulting policy shift to
Qwen3-4B-nonthinking reaches the 0.635 direct-RL level on AIME 2025. This
suggests that a small-model RL run can recover the useful RL direction even
when the target is a stronger 4B model.

\takeaway{Compared with running RL directly on the large model, first running RL
on the small model and then transferring to the large model has two advantages:
small-model RL is more stable, easier, and cheaper, while the following
\method{} stage provides fast transfer to the large model while preserving the performance advantage.}

\subsection{Sequential composition}
\label{sec:sequential-composition}

\begin{figure}[H]
\centering
\begin{minipage}[c]{0.63\linewidth}
\centering
\includegraphics[width=\linewidth]{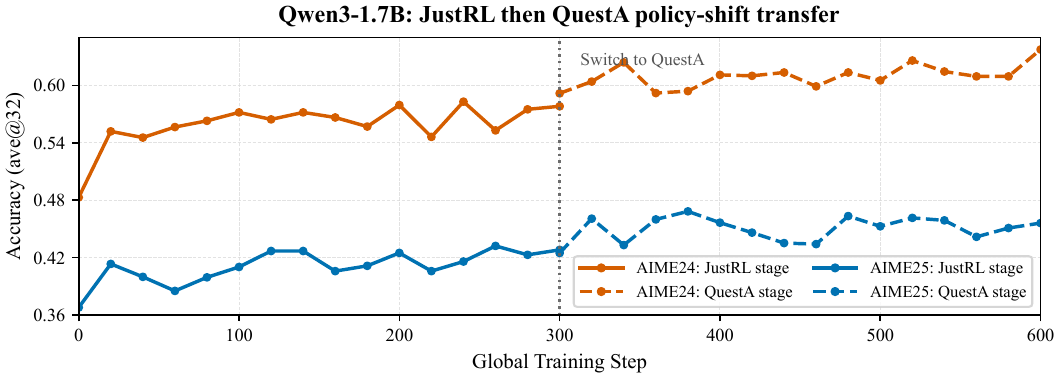}
\end{minipage}\hfill
\begin{minipage}[c]{0.33\linewidth}
\centering
\textbf{Score summary}\par\vspace{0.35em}
\scriptsize
\setlength{\tabcolsep}{2.3pt}
\renewcommand{\arraystretch}{1.08}
\begin{tabular}{lcc}
\toprule
Stage & AIME24 & AIME25 \\
\midrule
Initial & 48.3 & 36.8 \\
\rowcolor{cyan!12}
After JustRL & \textbf{58.3} \gain{+10.0} & \textbf{43.2} \gain{+6.4} \\
\rowcolor{cyan!12}
After QuestA & \textbf{63.8} \gain{+15.5} & \textbf{46.8} \gain{+10.0} \\
\bottomrule
\end{tabular}
\end{minipage}
\caption{Sequential policy-shift transfer into Qwen3-1.7B on AIME 2024 and
AIME 2025. Left: trajectories after aligning the second stage to global
steps 300--600. Right: endpoint scores on AIME 2024/2025. The first stage uses
the R1-Distill-1.5B $\rightarrow$ JustRL-1.5B signal; the second stage continues
from that checkpoint with the Nemotron-1.5B $\rightarrow$
QuestA-Nemotron-1.5B signal.}
\label{fig:sequential-justrl-questa-qwen3-aime24}
\end{figure}

This experiment asks whether two independently learned policy shifts can be
applied in sequence to the same student. The student is first trained with the
R1-Distill-1.5B $\rightarrow$ JustRL-1.5B signal and then continued with the
Nemotron-1.5B $\rightarrow$ QuestA-Nemotron-1.5B signal. Figure~\ref{fig:sequential-justrl-questa-qwen3-aime24}
shows the AIME 2024 and AIME 2025 trajectories after aligning the second stage's local
0--300 training steps to global steps 300--600. The first-stage endpoint and
the second-stage starting point are evaluated from separate samples, so the
small discontinuity at the stage boundary reflects evaluation variance.

\takeaway{Different RL training runs can learn different abilities, and
\method{} can sequentially compose these abilities into the same student.}

\section{Analysis and Training Dynamics}

The previous section showed \emph{that} \method{} transfers RL gains
weak-to-strong; here we analyze \emph{when} and \emph{why} it works, examining
three aspects of the transfer:
\begin{itemize}
    \item \textbf{What is transferred?} Whether \method{} needs the student to
    imitate the teacher's high-probability tokens, or whether it transfers even
    when teacher--student top-$k$ overlap stays low
    (Section~\ref{sec:cross-pattern}).
    \item \textbf{Does short-horizon training generalize?} Whether training on
    short rollouts changes only the supervised prefixes or shifts behavior across
    much longer responses (Section~\ref{sec:training-length}).
    \item \textbf{Why adapt the KL coefficient?} How the KL coefficient affects
    the reliability of the teacher-shift reward across teacher--student pairs
    (Section~\ref{sec:kl-coefficient}).
\end{itemize}

\subsection{Cross-pattern transfer without token-overlap imitation}
\label{sec:cross-pattern}

Prior work on standard OPD argues that teacher-student thinking-pattern
compatibility is a key success condition: successful OPD is reflected in
progressive alignment on high-probability tokens at student-visited states, with
a small shared top-$k$ token set carrying most of the probability
mass~\citep{li2026rethinkingopd}. This requirement is natural for raw teacher
imitation, because the student is asked to match the teacher's final
distribution. In that setting, increasing top-$k$ overlap is both a diagnostic
and a plausible carrier of the training signal.

We compute top-$k$ overlap using the same set-overlap metric as prior OPD work.
For a student policy $\pi_S$ and a comparison policy $\pi_C$ at state $s_t$, let
\begin{align}
    T_k^S(s_t) &= \operatorname{TopK}_{v}\ \pi_S(v\mid s_t), \\
    T_k^C(s_t) &= \operatorname{TopK}_{v}\ \pi_C(v\mid s_t).
\end{align}
The per-state overlap ratio is
\begin{align}
    \operatorname{Overlap}_k(S,C;s_t)
    =
    \frac{\left|T_k^S(s_t)\cap T_k^C(s_t)\right|}{k}.
\end{align}
For \method{}, we report this metric against both the post-RL teacher
($C=T$) and the teacher reference ($C=T_{\mathrm{ref}}$) on the same
student-visited states.

\method{} changes the object being transferred. Instead of matching the
post-RL teacher distribution, it evaluates the teacher/reference log-ratio on
student-visited candidate tokens. This signal can rank actions within the
student's own support, so useful transfer need not require the student to enter
the teacher's high-overlap token regime.

\begin{figure}[htbp]
\centering
\begin{minipage}[t]{0.49\linewidth}
\vspace{0pt}
\centering
\includegraphics[width=\linewidth]{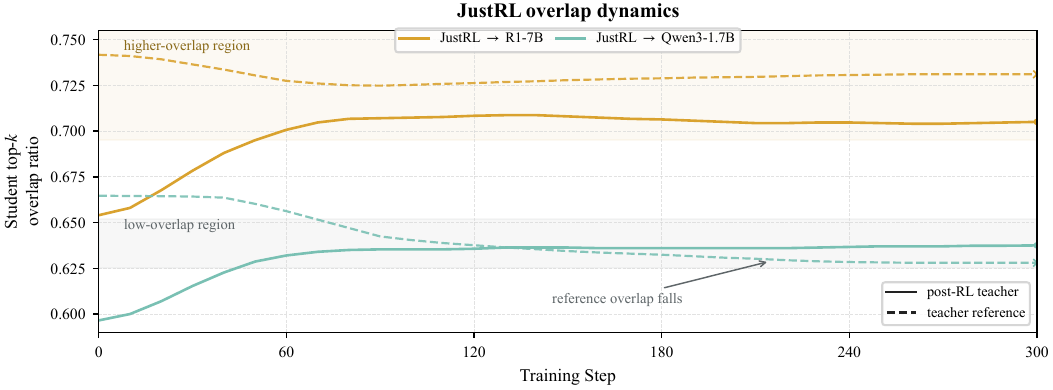}
\end{minipage}
\hfill
\begin{minipage}[t]{0.49\linewidth}
\vspace{0pt}
\centering
\includegraphics[width=\linewidth]{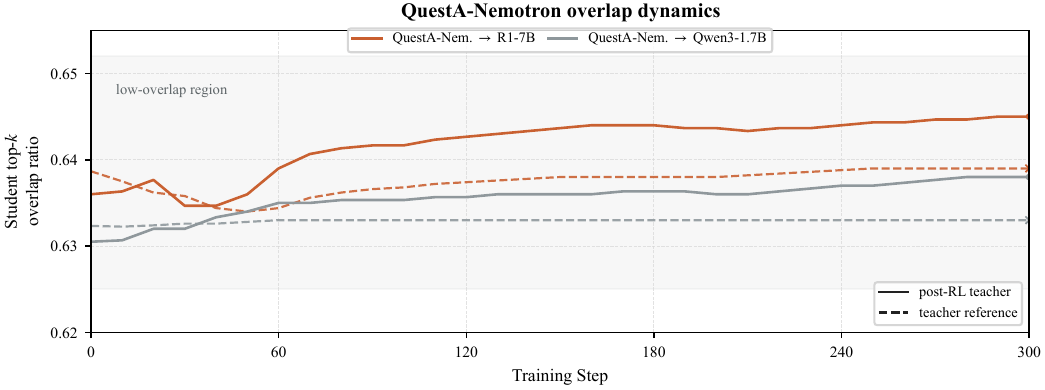}
\end{minipage}
\caption{Teacher--student top-$k$ overlap during \method{} training. Left:
R1-Distill-1.5B $\rightarrow$ JustRL-1.5B teacher pair. Right:
Nemotron-1.5B $\rightarrow$ QuestA-Nemotron-1.5B teacher pair. Solid curves
measure overlap with the post-RL teacher, and dashed curves measure overlap
with the teacher reference. The pattern-aligned R1-Distill transfer enters a
higher-overlap regime, while the cross-pattern transfers remain lower and do
not become imitation of the post-RL teacher.}
\label{fig:justrl-overlap-cross-pattern}
\label{fig:questa-overlap-cross-pattern}
\end{figure}

Figure~\ref{fig:justrl-overlap-cross-pattern} separates the pattern-aligned
case from the cross-pattern cases. The aligned transfer follows the standard
OPD intuition: the student moves into a higher-overlap regime with the post-RL
teacher. The cross-pattern transfers behave differently. Their overlap with the
post-RL teacher remains lower, and their overlap with the teacher reference does
not show a compensating rise. Thus the gains in
Figure~\ref{fig:transfer-generalization} are not explained by progressive
imitation of either teacher checkpoint. They are better explained by using the
RL-induced direction from reference to post-RL teacher on the student's own
visited states.

We next check whether this low-overlap transfer is a degenerate sharpening
effect. If \method{} simply collapsed the actor or forced a trivial low-entropy
distribution, the overlap diagnostic would be less informative.

\begin{figure}[htbp]
\centering
\includegraphics[width=\linewidth]{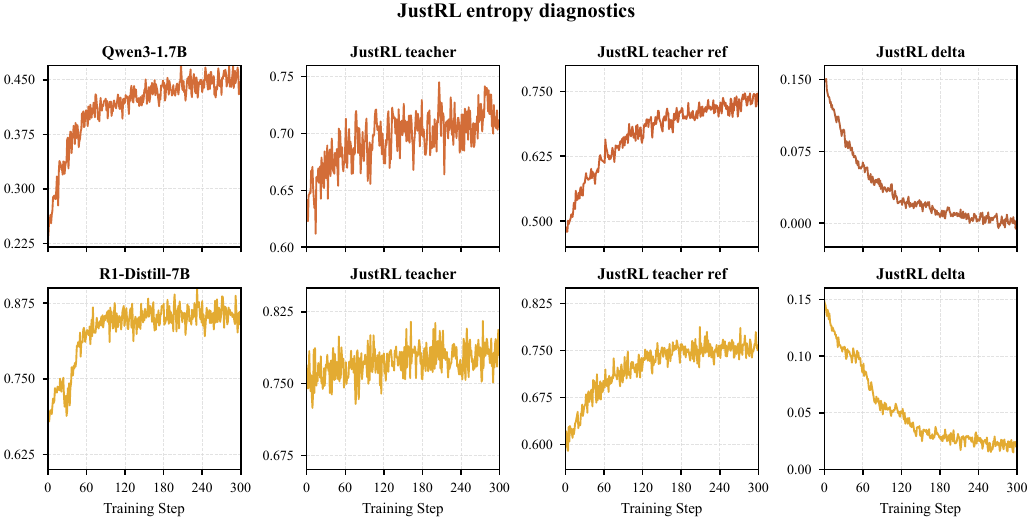}
\caption{Entropy diagnostics for R1-Distill-1.5B $\rightarrow$ JustRL-1.5B
policy-shift transfer. \textbf{Top row:} transfer into Qwen3-1.7B.
\textbf{Bottom row:} transfer into R1-Distill-7B. Each row shows student
entropy, post-RL teacher entropy, teacher-reference entropy, and teacher entropy
minus reference entropy. Actor entropy does not collapse, while the
teacher/reference entropy gap narrows over training.}
\label{fig:qwen3-entropy}
\end{figure}

Figure~\ref{fig:qwen3-entropy} shows that the actor entropy remains controlled
rather than collapsing. At the same time, the teacher/reference entropy gap
narrows, indicating that training changes how the student samples regions where
the teacher's RL-induced shift is evaluated. This supports the same mechanism
as the overlap result: \method{} transfers a local policy-shift signal within
the student's distribution, not a full teacher policy.

\takeaway{\method{} does not require progressive token-overlap
imitation. It can transfer the teacher's RL-induced direction on
student-visited tokens while keeping the actor distribution controlled.}

\subsection{Short-horizon training changes longer-rollout behavior}
\label{sec:training-length}

Recent work on on-policy prefix distillation shows that full-length supervision
is not always necessary for reasoning transfer: student rollouts can be
truncated to short prefixes, while the resulting model is still evaluated with
long generations at test time~\citep{zhang2026prefixopd}. Following this view,
we do not treat the training horizon merely as a finite rollout length. We
deliberately use a short 2k-token response horizon for \method{} training, and
ask whether the resulting policy change generalizes beyond the directly
supervised prefix.

A natural concern is that such short-horizon training may only change the early
prefixes that receive direct supervision, leaving later reasoning behavior
unchanged. We test this by evaluating trained actors on fixed long rollouts and
measuring whether the actor's likely next tokens move toward the post-RL teacher
relative to the teacher reference.

\begin{figure}[htbp]
\centering
\includegraphics[width=\linewidth]{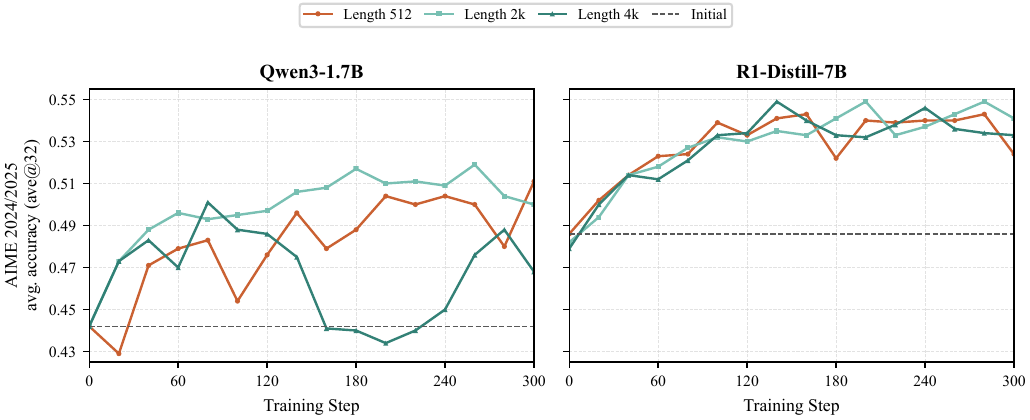}
\caption{Response-length sweep for R1-Distill-1.5B $\rightarrow$ JustRL-1.5B
transfer with fixed KL coefficient $1$. We report the average of AIME 2024 and
AIME 2025 validation accuracy (ave@32) during training for Qwen3-1.7B and
R1-Distill-7B students. The 2k setting gives stable validation behavior across
the two students, while shorter or longer rollouts do not consistently improve
the validation curves.}
\label{fig:justrl-length-sweep}
\end{figure}

For each response prefix
$x_{\leq t}$, we take the actor's top-16 token set $K_t$ and compute the
actor-probability-weighted teacher gap
\begin{equation}
g_t =
\sum_{a \in K_t} \pi_{\mathrm{actor}}(a \mid x_{\leq t})
\left[
\log \pi_{\mathrm{JustRL}}(a \mid x_{\leq t})
-
\log \pi_{\mathrm{R1}}(a \mid x_{\leq t})
\right].
\end{equation}
The plotted quantity is the prefix cumulative gap
\begin{equation}
G_T = \sum_{t=1}^{T} g_t .
\end{equation}
Larger $G_T$ means that, along the actor's own likely tokens, the JustRL teacher
assigns higher weighted log-probability than the R1-Distill-1.5B teacher; smaller
$G_T$ means that the actor remains closer to R1-Distill-1.5B.

\begin{figure}[htbp]
\centering
\includegraphics[width=\linewidth]{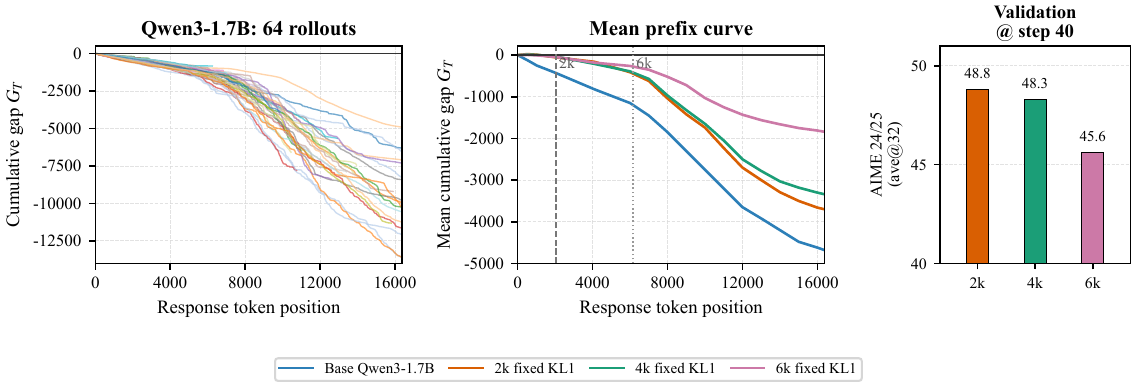}
\caption{\textbf{Short-horizon \method{} training changes behavior beyond the
supervised prefix.} On a fixed set of 64 long
rollouts we track the cumulative gap $G_T=\sum_{t\le T} g_t$, where $g_t$
weights, over the actor's top-16 tokens at position $t$, the log-probability the
post-RL teacher (JustRL) assigns minus that of the reference (R1-Distill-1.5B);
higher $G_T$ means the actor's likely tokens look more like the post-RL teacher, while lower means the actor remains closer to the reference.
\textbf{Left:} the 64 per-rollout trajectories for the untrained Qwen3-1.7B
actor all drift negative---its long rollouts are reference-like, and this is not
driven by a single outlier. \textbf{Middle:} the mean $G_T$ for the base actor
and for actors trained with 2k/4k/6k response length (40 steps, fixed
KL${=}1$); every trained actor sits above the base across the full $\sim$16k
positions, and the 2k actor shifts well past its 2k training horizon (dashed
line). \textbf{Right:} AIME 2024/2025 (ave@32) at the same 40-step
checkpoint---the 2k setting validates best.}
\label{fig:response-length-prefix-gap}
\end{figure}

Figure~\ref{fig:response-length-prefix-gap} shows that the effect of short
\method{} training is not confined to the supervised prefix range: after 40
training steps with a 2k response length, the actor moves toward the
teacher-shift direction across much longer fixed rollouts, so the short horizon
already captures the transferable signal. Training with a 6k response length
shifts the actor even further in this diagnostic (middle panel), yet it
validates worst (right panel, $45.6$ versus $48.8$ for 2k). One plausible
explanation is that the extra movement comes from late prefixes, where the
small teacher pair is increasingly off-distribution and the teacher/reference
log-ratio is large but unreliable
(Section~\ref{sec:kl-coefficient}). A 6k horizon may over-drive the actor on
this unreliable late-prefix signal without improving the answer, whereas a 2k
horizon avoids ingesting it while still capturing---and generalizing---the
reliable early-prefix direction.

\takeaway{Short-horizon \method{} training can generalize changes in thinking patterns to
longer rollouts. A moderate response length avoids the noise and instability of
very long responses while capturing more useful signal than overly short
training horizons.}

\subsection{KL controls teacher-shift reward reliability}
\label{sec:kl-coefficient}

The teacher/reference log-ratio acts as a dense reward, but this reward is only
reliable on states where the teacher shift is meaningful for the student. This
makes KL more than a regularization coefficient: it controls which rollout
states the student visits, and therefore whether the dense reward remains
reliable. If KL is too weak, the student can drift into regions where both
teacher and reference assign low probability and the log-ratio becomes a noisy
training target. If KL is too strong, the student cannot follow the teacher's
policy shift. We test this by sweeping fixed KL coefficients under the same
2k-response setting and comparing them with an adaptive-KL run.

\begin{figure}[htbp]
\centering
\includegraphics[width=\linewidth]{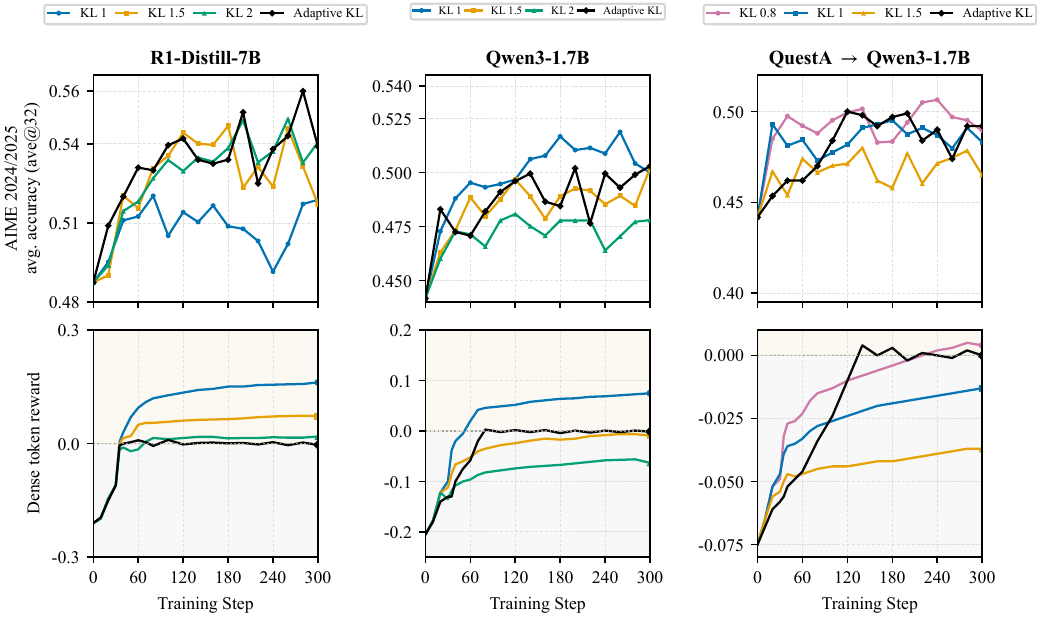}
\caption{\textbf{The best KL coefficient is pair-dependent, and adaptive KL
pulls the dense token reward toward a balanced regime.} 2k-response runs
across fixed KL coefficients, with adaptive KL as the black curve. \textbf{Top
row:} AIME 2024/2025 validation accuracy (ave@32); \textbf{bottom row:} dense
token reward. The best fixed coefficient differs across teacher-student
pairs, and a larger reward does not imply better validation. Adaptive KL
instead pulls the dense token reward toward zero after an initial correction, keeping
the student from simply maximizing the dense teacher/reference reward.}
\label{fig:justrl-kl-sweep-score-critic}
\end{figure}

Figure~\ref{fig:justrl-kl-sweep-score-critic} shows that the best fixed KL
value is not universal: different teacher-student pairs prefer different
constraint strengths. The dense token reward also does not provide a
pair-independent rule; a larger positive value can coincide with worse
validation in one setting but track the best validation curve in another. Thus
the dense reward should not be treated as a scalar objective to maximize
independently of the rollout distribution. The useful regime depends on the
teacher-student pair and on where the student currently samples.

The adaptive-KL runs provide a complementary diagnostic. After an initial
correction phase, their dense token rewards move toward zero rather than
remaining strongly positive or negative. This behavior is consistent with the
intended role of adaptive KL: keep the student in a region where the
teacher/reference comparison remains informative, instead of letting the student
either ignore the teacher shift or exploit an unreliable local signal. A
persistently large reward can indicate that the student is sampling tokens on
which the post-RL teacher and its reference strongly disagree---often because it
has drifted away from both models' support, where the log-ratio is large in
magnitude but no longer a trustworthy improvement direction. A near-zero batch
mean suggests a more balanced regime, where local per-token shifts can still
rank actions without a global drift.

\takeaway{The dense teacher/reference reward should be treated as
rollout-distribution dependent rather than optimized in isolation. Fixed-KL
sweeps show that the right constraint is pair-dependent, while adaptive KL
empirically pulls the mean reward toward a more balanced regime.}

\section{Related Work}

\paragraph{On-policy distillation of reasoning models.}
Reinforcement learning with verifiable rewards now produces strong reasoning
models~\citep{deepseekr1,shao2024deepseekmath,schulman2017ppo,k1p5,qwq,openreasonerzero,dapo,justrl,polaris,questa,he2025skywork,he2025deepmath,he2026far},
a paradigm now standard in frontier
systems~\citep{qwen3technicalreport,deepseekv4,glm52,openmathreasoning,guha2025openthoughts,li2024numinamath}.
A standard way to spread such gains to other models is knowledge distillation,
from classic and sequence-level distillation to its modern variants and scaling
behavior~\citep{hinton2015distilling,skd,sanh2019distilbert,jiao2020tinybert,wang2020minilm,mirzadeh2020improved,cho2019efficacy,busbridge2025distillation},
with a large body of work on divergences and estimators for distilling
LLMs~\citep{minillm,wen2023fdistill,gu2024revisitkd,wu2024rethinkkl,ko2024distillm,ko2025distillm2,miniplm,zhou2024distillspec,xu2024speckd}.
On-policy distillation (OPD), where the student is trained on its own sampled
states under teacher supervision, has become the dominant variant for reasoning
models and is the subject of rapid recent
work~\citep{gkd,lu2025onpolicydistillation,li2026rethinkingopd,opdsurvey,fu2026revisiting,jin2026entropy,jang2026stable,li2026unifying,ye2026policy,ye2026online,zhao2026self,zhao2026selfdistillationmultitokenprediction,rlselfdistill,shenfeld2026self,zhang2026prefixopd,paced,ko2026scaling,ko2026scaling,kim2026does,zhang2025blackboxopd,xiao2026mimo},
including hybrids that couple distillation with
RL~\citep{rlkd,kdrl,rakd,aligndistil,rlselfdistill,yang2026self}.
Analyses tie OPD success to teacher--student top-$k$ overlap on student-visited
states~\citep{li2026rethinkingopd}, note that small students struggle to copy
much stronger reasoners~\citep{li2025small}, and reinterpret OPD as dense
KL-constrained RL that rescales the reward to extrapolate beyond the
teacher~\citep{gopd}. All of these imitate, or push past, the teacher's final
policy; when the teacher is a small post-RL model weaker than the target, this
also imports its capacity ceiling. \method{} instead distills only the teacher's
RL-induced \emph{policy shift}
$\log\pi_T-\log\pi_{T_{\mathrm{ref}}}$, discarding the teacher's absolute policy.

\paragraph{Weak-to-strong generalization.}
Weak-to-strong generalization asks whether a weak supervisor can elicit
capabilities from a stronger model~\citep{burns2023weakstrong,superalignment},
building on a long line of learning from imperfect supervision---semi-supervised
learning and learning from noisy
labels~\citep{grandvalet2004semi,dai2015semi,laine2016temporal,athiwaratkun2018there,han2018co,berthelot2019mixmatch,li2020dividemix,chen2020big,chen2020self,chen2022debiased}.
It is central to scalable oversight when reliable human or stronger-model
supervision is scarce~\citep{christiano2018supervising,leike2018scalable,irving2018debate,bowman2022scalable,bowman2022alignment,bai2022constitutional},
and is related to eliciting latent knowledge and to easy-to-hard
generalization~\citep{christiano2022eliciting,burns2022discovering,schwarzschild2021can,schwarzschild2021datasets}.
Recent work studies its theory and extends it to reasoning and automated
alignment~\citep{survey_autoalign,bansal2025revisiting_w2s,w2s_fdivergence,guo2024improving_w2s,w2s_reasoning}.
Most approaches still supervise with the weak model's \emph{labels or
distribution}, which caps the student near the supervisor; self-rewarding schemes
bootstrap a model with its own judgments~\citep{yuan2024self}, and, closest to
us, weak-to-strong preference optimization reuses a weak aligned model's implicit
reward to steer a stronger one~\citep{wspo}. We share the goal of eliciting
rather than imitating, but our supervisor is a \emph{pair} of small RL
checkpoints: we transfer the reward implied by their difference, and show it
improves students that already exceed the post-RL teacher.

\paragraph{Implicit rewards and reusing trained models.}
\method{} rests on the policy-as-reward identity: under KL-regularized RL the
policy/reference log-ratio recovers the reward up to a constant, which Direct
Preference Optimization and its variants use to fit a policy directly from
preferences without an explicit reward
model~\citep{rafailov2023direct,azar2024ipo,ethayarajh2024kto,meng2024simpo}.
We use this identity in reverse---reading a dense implicit reward off a post-RL
checkpoint and its reference---which connects to dense and process rewards for
reasoning~\citep{lightman2023let,mathshepherd,cui2025process} and to per-token
credit assignment in RL~\citep{kazemnejad2024vineppo,creditassignsurvey}, while
avoiding reward over-optimization from an explicit
model~\citep{gao2023overopt}. A complementary line reuses trained models in
\emph{weight} space---task arithmetic, model merging, and proxy
tuning~\citep{ilharco2023task,wortsman2022soups,proxytuning}. \method{} differs
on both counts: the signal is a behavioral log-ratio rather than a weight delta,
and it is applied on the student's own on-policy states, so it transfers across
model families and scales.

\section{Conclusion and Limitations}

\method{} shows that the policy shift learned by a small RL teacher is a useful
dense reward for weak-to-strong generalization. The transferable object is not
the post-RL teacher's policy, but its log-ratio against its own pre-RL reference,
evaluated on student-visited tokens; this is why a smaller, weaker teacher can
still improve stronger students. Empirically, this direction transfers across
teacher pairs and student families, outperforms step-matched direct RL at a
fraction of the compute, and composes across teachers. It reframes RL outcomes as
reusable improvement signals rather than final models to imitate.
\textbf{Limitations.} The signal is conditional: \method{} can fail when the
teacher/reference improvement is not meaningful on student-visited states, and
the best response length and KL strength remain teacher--student dependent.

\newpage

\bibliography{references}

\begin{thebibliography}{105}
\providecommand{\natexlab}[1]{#1}
\providecommand{\url}[1]{\texttt{#1}}
\expandafter\ifx\csname urlstyle\endcsname\relax
  \providecommand{\doi}[1]{doi: #1}\else
  \providecommand{\doi}{doi: \begingroup \urlstyle{rm}\Url}\fi

\bibitem[{DeepSeek-AI} et~al.(2025){DeepSeek-AI}, Guo, Yang, Zhang, Song, Wang,
  Zhu, Xu, Zhang, Ma, Bi, et~al.]{deepseekr1}
{DeepSeek-AI}, Daya Guo, Dejian Yang, Haowei Zhang, Junxiao Song, Peiyi Wang,
  Qihao Zhu, Runxin Xu, Ruoyu Zhang, Shirong Ma, Xiao Bi, et~al.
\newblock {DeepSeek-R1}: Incentivizing reasoning capability in {LLMs} via
  reinforcement learning.
\newblock \emph{Nature}, 645:\penalty0 633--638, 2025.

\bibitem[He et~al.(2025{\natexlab{a}})He, Qu, Liu, Chen, Zuo, Qian, Zhang,
  Chen, Xiao, Cui, Ding, and Liu]{justrl}
Bingxiang He, Zekai Qu, Zeyuan Liu, Yinghao Chen, Yuxin Zuo, Cheng Qian, Kaiyan
  Zhang, Weize Chen, Chaojun Xiao, Ganqu Cui, Ning Ding, and Zhiyuan Liu.
\newblock {JustRL}: Scaling a 1.5{B} {LLM} with a simple {RL} recipe.
\newblock \emph{arXiv preprint arXiv:2512.16649}, 2025{\natexlab{a}}.

\bibitem[Yang et~al.(2025)Yang, Li, Yang, Zhang, Hui, Zheng, Yu, Gao, Huang,
  Lv, et~al.]{qwen3technicalreport}
An~Yang, Anfeng Li, Baosong Yang, Beichen Zhang, Binyuan Hui, Bo~Zheng, Bowen
  Yu, Chang Gao, Chengen Huang, Chenxu Lv, et~al.
\newblock {Qwen3} technical report.
\newblock \emph{arXiv preprint arXiv:2505.09388}, 2025.

\bibitem[An et~al.(2025)]{polaris}
Chenxin An et~al.
\newblock {POLARIS}: A post-training recipe for scaling reinforcement learning
  on reasoning models.
\newblock \url{https://github.com/ChenxinAn-fdu/POLARIS}, 2025.

\bibitem[Li et~al.(2026{\natexlab{a}})Li, Zuo, He, Zhang, Xiao, Qian, Yu, Gao,
  Yang, Liu, and Ding]{li2026rethinkingopd}
Yaxuan Li, Yuxin Zuo, Bingxiang He, Jinqian Zhang, Chaojun Xiao, Cheng Qian,
  Tianyu Yu, Huan-ang Gao, Wenkai Yang, Zhiyuan Liu, and Ning Ding.
\newblock Rethinking on-policy distillation of large language models:
  Phenomenology, mechanism, and recipe.
\newblock \emph{arXiv preprint arXiv:2604.13016}, 2026{\natexlab{a}}.

\bibitem[Rafailov et~al.(2023)Rafailov, Sharma, Mitchell, Ermon, Manning, and
  Finn]{rafailov2023direct}
Rafael Rafailov, Archit Sharma, Eric Mitchell, Stefano Ermon, Christopher~D.
  Manning, and Chelsea Finn.
\newblock Direct preference optimization: Your language model is secretly a
  reward model.
\newblock \emph{arXiv preprint arXiv:2305.18290}, 2023.

\bibitem[Li et~al.(2025{\natexlab{a}})Li, Lin, Lu, Wen, Yang, Gao, Wu, and
  Zhang]{questa}
Jiazheng Li, Hongzhou Lin, Hong Lu, Kaiyue Wen, Zaiwen Yang, Jiaxuan Gao,
  Yi~Wu, and Jingzhao Zhang.
\newblock {QuestA}: Expanding reasoning capacity in {LLMs} via question
  augmentation.
\newblock \emph{arXiv preprint arXiv:2507.13266}, 2025{\natexlab{a}}.

\bibitem[Zhang et~al.(2026{\natexlab{a}})Zhang, Yang, Janghorbani, Han,
  Ressler~II, Qian, Lyng, Batra, and Tillman]{zhang2026prefixopd}
Dongxu Zhang, Zhichao Yang, Sepehr Janghorbani, Jun Han, Andrew Ressler~II,
  Qian Qian, Gregory~D. Lyng, Sanjit~Singh Batra, and Robert~E. Tillman.
\newblock Fast and effective on-policy distillation from reasoning prefixes.
\newblock In \emph{Findings of the Association for Computational Linguistics:
  ACL 2026}, pages 25553--25569, 2026{\natexlab{a}}.

\bibitem[Shao et~al.(2024)Shao, Wang, Zhu, Xu, Song, Bi, Zhang, Zhang, Li, Wu,
  and Guo]{shao2024deepseekmath}
Zhihong Shao, Peiyi Wang, Qihao Zhu, Runxin Xu, Junxiao Song, Xiao Bi, Haowei
  Zhang, Mingchuan Zhang, Y.~K. Li, Y.~Wu, and Daya Guo.
\newblock {DeepSeekMath}: Pushing the limits of mathematical reasoning in open
  language models.
\newblock \emph{arXiv preprint arXiv:2402.03300}, 2024.

\bibitem[Schulman et~al.(2017)Schulman, Wolski, Dhariwal, Radford, and
  Klimov]{schulman2017ppo}
John Schulman, Filip Wolski, Prafulla Dhariwal, Alec Radford, and Oleg Klimov.
\newblock Proximal policy optimization algorithms.
\newblock \emph{arXiv preprint arXiv:1707.06347}, 2017.

\bibitem[{Kimi Team}(2025)]{k1p5}
{Kimi Team}.
\newblock Kimi k1.5: Scaling reinforcement learning with llms.
\newblock \emph{arXiv preprint arXiv:2501.12599}, 2025.

\bibitem[{Qwen Team}(2025)]{qwq}
{Qwen Team}.
\newblock {QwQ-32B}: Embracing the power of reinforcement learning.
\newblock \url{https://qwenlm.github.io/blog/qwq-32b/}, 2025.

\bibitem[Hu et~al.(2025)Hu, Zhang, Han, Jiang, Zhang, and
  Shum]{openreasonerzero}
Jingcheng Hu, Yinmin Zhang, Qi~Han, Daxin Jiang, Xiangyu Zhang, and Heung-Yeung
  Shum.
\newblock Open-reasoner-zero: An open source approach to scaling reinforcement
  learning on the base model.
\newblock \url{https://github.com/Open-Reasoner-Zero/Open-Reasoner-Zero}, 2025.

\bibitem[Yu et~al.(2025)Yu, Zhang, Zhu, Yuan, Zuo, Yue, Fan, Liu, Liu, Liu,
  et~al.]{dapo}
Qiying Yu, Zheng Zhang, Ruofei Zhu, Yufeng Yuan, Xiaochen Zuo, Yu~Yue, Tiantian
  Fan, Gaohong Liu, Lingjun Liu, Xin Liu, et~al.
\newblock {DAPO}: An open-source llm reinforcement learning system at scale.
\newblock \emph{arXiv preprint arXiv:2503.14476}, 2025.

\bibitem[He et~al.(2025{\natexlab{b}})He, Liu, Liu, Yan, Wang, Cheng, Zhang,
  Zhang, Xu, Shen, et~al.]{he2025skywork}
Jujie He, Jiacai Liu, Chris~Yuhao Liu, Rui Yan, Chaojie Wang, Peng Cheng,
  Xiaoyu Zhang, Fuxiang Zhang, Jiacheng Xu, Wei Shen, et~al.
\newblock Skywork open reasoner 1 technical report.
\newblock \emph{arXiv preprint arXiv:2505.22312}, 2025{\natexlab{b}}.

\bibitem[He et~al.(2025{\natexlab{c}})He, Liang, Xu, Liu, Chen, Wang, Song, Yu,
  Liang, Wang, et~al.]{he2025deepmath}
Zhiwei He, Tian Liang, Jiahao Xu, Qiuzhi Liu, Xingyu Chen, Yue Wang, Linfeng
  Song, Dian Yu, Zhenwen Liang, Wenxuan Wang, et~al.
\newblock Deepmath-103k: A large-scale, challenging, decontaminated, and
  verifiable mathematical dataset for advancing reasoning.
\newblock \emph{arXiv preprint arXiv:2504.11456}, 2025{\natexlab{c}}.

\bibitem[He et~al.(2026)He, Zuo, Liu, Zhao, Fu, Yang, Qian, Zhang, Fan, Cui,
  et~al.]{he2026far}
Bingxiang He, Yuxin Zuo, Zeyuan Liu, Shangziqi Zhao, Zixuan Fu, Junlin Yang,
  Cheng Qian, Kaiyan Zhang, Yuchen Fan, Ganqu Cui, et~al.
\newblock How far can unsupervised rlvr scale llm training?
\newblock \emph{arXiv preprint arXiv:2603.08660}, 2026.

\bibitem[{DeepSeek-AI}(2026)]{deepseekv4}
{DeepSeek-AI}.
\newblock {DeepSeek V4} preview release.
\newblock \url{https://api-docs.deepseek.com/news/news260424}, 2026.

\bibitem[{Z.ai}(2026)]{glm52}
{Z.ai}.
\newblock {GLM-5.2}: Built for long-horizon tasks.
\newblock \url{https://z.ai/blog/glm-5.2}, 2026.

\bibitem[Moshkov et~al.(2025)Moshkov, Hanley, Sorokin, Toshniwal, Henkel,
  Schifferer, Du, and Gitman]{openmathreasoning}
Ivan Moshkov, Darragh Hanley, Ivan Sorokin, Shubham Toshniwal, Christof Henkel,
  Benedikt Schifferer, Wei Du, and Igor Gitman.
\newblock {AIMO-2} winning solution: Building state-of-the-art mathematical
  reasoning models with {OpenMathReasoning} dataset.
\newblock \emph{arXiv preprint arXiv:2504.16891}, 2025.

\bibitem[Guha et~al.(2025)Guha, Marten, Keh, Raoof, Smyrnis, Bansal, Nezhurina,
  Mercat, Vu, Sprague, et~al.]{guha2025openthoughts}
Etash Guha, Ryan Marten, Sedrick Keh, Negin Raoof, Georgios Smyrnis, Hritik
  Bansal, Marianna Nezhurina, Jean Mercat, Trung Vu, Zayne Sprague, et~al.
\newblock Openthoughts: Data recipes for reasoning models.
\newblock \emph{arXiv preprint arXiv:2506.04178}, 2025.

\bibitem[Li et~al.(2024)Li, Beeching, Tunstall, Lipkin, Soletskyi, Huang,
  Rasul, Yu, Jiang, Shen, et~al.]{li2024numinamath}
Jia Li, Edward Beeching, Lewis Tunstall, Ben Lipkin, Roman Soletskyi, Shengyi
  Huang, Kashif Rasul, Longhui Yu, Albert~Q Jiang, Ziju Shen, et~al.
\newblock Numinamath: The largest public dataset in ai4maths with 860k pairs of
  competition math problems and solutions.
\newblock 2024.

\bibitem[Hinton et~al.(2015)Hinton, Vinyals, and Dean]{hinton2015distilling}
Geoffrey Hinton, Oriol Vinyals, and Jeff Dean.
\newblock Distilling the knowledge in a neural network.
\newblock \emph{arXiv preprint arXiv:1503.02531}, 2015.

\bibitem[Kim and Rush(2016)]{skd}
Yoon Kim and Alexander~M Rush.
\newblock Sequence-level knowledge distillation.
\newblock In \emph{Proceedings of EMNLP}, 2016.

\bibitem[Sanh et~al.(2019)Sanh, Debut, Chaumond, and Wolf]{sanh2019distilbert}
Victor Sanh, Lysandre Debut, Julien Chaumond, and Thomas Wolf.
\newblock Distilbert, a distilled version of bert: smaller, faster, cheaper and
  lighter.
\newblock \emph{arXiv preprint arXiv:1910.01108}, 2019.

\bibitem[Jiao et~al.(2020)Jiao, Yin, Shang, Jiang, Chen, Li, Wang, and
  Liu]{jiao2020tinybert}
Xiaoqi Jiao, Yichun Yin, Lifeng Shang, Xin Jiang, Xiao Chen, Linlin Li, Fang
  Wang, and Qun Liu.
\newblock Tinybert: Distilling bert for natural language understanding.
\newblock 2020.

\bibitem[Wang et~al.(2020)Wang, Wei, Dong, Bao, Yang, and Zhou]{wang2020minilm}
Wenhui Wang, Furu Wei, Li~Dong, Hangbo Bao, Nan Yang, and Ming Zhou.
\newblock Minilm: Deep self-attention distillation for task-agnostic
  compression of pre-trained transformers.
\newblock 2020.

\bibitem[Mirzadeh et~al.(2020)Mirzadeh, Farajtabar, Li, Levine, Matsukawa, and
  Ghasemzadeh]{mirzadeh2020improved}
Seyed~Iman Mirzadeh, Mehrdad Farajtabar, Ang Li, Nir Levine, Akihiro Matsukawa,
  and Hassan Ghasemzadeh.
\newblock Improved knowledge distillation via teacher assistant.
\newblock 2020.

\bibitem[Cho and Hariharan(2019)]{cho2019efficacy}
Jang~Hyun Cho and Bharath Hariharan.
\newblock On the efficacy of knowledge distillation.
\newblock 2019.

\bibitem[Busbridge et~al.(2025)Busbridge, Shidani, Weers, Ramapuram, Littwin,
  and Webb]{busbridge2025distillation}
Dan Busbridge, Amitis Shidani, Floris Weers, Jason Ramapuram, Etai Littwin, and
  Russ Webb.
\newblock Distillation scaling laws.
\newblock \emph{arXiv preprint arXiv:2502.08606}, 2025.

\bibitem[Gu et~al.(2023)Gu, Dong, Wei, and Huang]{minillm}
Yuxian Gu, Li~Dong, Furu Wei, and Minlie Huang.
\newblock {MiniLLM}: Knowledge distillation of large language models.
\newblock \emph{arXiv preprint arXiv:2306.08543}, 2023.

\bibitem[Wen et~al.(2023)Wen, Li, Du, and Mou]{wen2023fdistill}
Yuqiao Wen, Zichao Li, Wenyu Du, and Lili Mou.
\newblock {f-Divergence Minimization for Sequence-Level Knowledge
  Distillation}.
\newblock \emph{Proceedings of ACL}, 2023.
\newblock URL \url{https://arxiv.org/abs/2307.15190}.

\bibitem[Zhong et~al.(2024)Zhong, Ding, Shen, Liu, Du, and
  Tao]{gu2024revisitkd}
Qihuang Zhong, Liang Ding, Li~Shen, Juhua Liu, Bo~Du, and Dacheng Tao.
\newblock {Revisiting Knowledge Distillation for Autoregressive Language
  Models}.
\newblock \emph{Proceedings of ACL}, 2024.
\newblock URL \url{https://arxiv.org/abs/2402.11890}.

\bibitem[Wu et~al.(2025)Wu, Tao, Wang, Yang, Zhao, and Wong]{wu2024rethinkkl}
Taiqiang Wu, Chaofan Tao, Jiahao Wang, Runming Yang, Zhe Zhao, and Ngai Wong.
\newblock {Rethinking Kullback-Leibler Divergence in Knowledge Distillation for
  Large Language Models}.
\newblock \emph{Proceedings of COLING}, 2025.
\newblock URL \url{https://arxiv.org/abs/2404.02657}.

\bibitem[Ko et~al.(2024)Ko, Kim, Chen, and Yun]{ko2024distillm}
Jongwoo Ko, Sungnyun Kim, Tianyi Chen, and Se-Young Yun.
\newblock {DistiLLM: Towards Streamlined Distillation for Large Language
  Models}.
\newblock \emph{Proceedings of ICML}, 2024.
\newblock URL \url{https://arxiv.org/abs/2402.03898}.

\bibitem[Ko et~al.(2025)Ko, Chen, Kim, Ding, Liang, Zharkov, and
  Yun]{ko2025distillm2}
Jongwoo Ko, Tianyi Chen, Sungnyun Kim, Tianyu Ding, Luming Liang, Ilya Zharkov,
  and Se-Young Yun.
\newblock {DistiLLM-2: A Contrastive Approach Boosts the Distillation of LLMs}.
\newblock In \emph{Proceedings of ICML}, 2025.
\newblock URL \url{https://arxiv.org/abs/2503.07067}.

\bibitem[Gu et~al.(2025)Gu, Zhou, Meng, Zhou, and Huang]{miniplm}
Yuxian Gu, Hao Zhou, Fandong Meng, Jie Zhou, and Minlie Huang.
\newblock {MiniPLM: Knowledge Distillation for Pre-Training Language Models}.
\newblock \emph{Proceedings of ICLR}, 2025.
\newblock URL \url{https://arxiv.org/abs/2410.17215}.

\bibitem[Zhou et~al.(2024)Zhou, Lyu, Rawat, Menon, Rostamizadeh, Kumar, Kagy,
  and Agarwal]{zhou2024distillspec}
Yongchao Zhou, Kaifeng Lyu, Ankit~Singh Rawat, Aditya~Krishna Menon, Afshin
  Rostamizadeh, Sanjiv Kumar, Jean-Fran\c{c}ois Kagy, and Rishabh Agarwal.
\newblock {DistillSpec: Improving Speculative Decoding via Knowledge
  Distillation}.
\newblock \emph{Proceedings of ICLR}, 2024.
\newblock URL \url{https://arxiv.org/abs/2310.08461}.

\bibitem[Xu et~al.(2025{\natexlab{a}})Xu, Han, Wang, Le, Madeka, Li, Wang,
  Agarwal, Lee, and Pfister]{xu2024speckd}
Wenda Xu, Rujun Han, Zifeng Wang, Long~T. Le, Dhruv Madeka, Lei Li,
  William~Yang Wang, Rishabh Agarwal, Chen-Yu Lee, and Tomas Pfister.
\newblock {Speculative Knowledge Distillation: Bridging the Teacher-Student Gap
  Through Interleaved Sampling}.
\newblock \emph{Proceedings of ICLR}, 2025{\natexlab{a}}.
\newblock URL \url{https://arxiv.org/abs/2410.11325}.

\bibitem[Agarwal et~al.(2023)Agarwal, Vieillard, Zhou, Stanczyk, Ramos, Geist,
  and Bachem]{gkd}
Rishabh Agarwal, Nino Vieillard, Yongchao Zhou, Piotr Stanczyk, Sabela Ramos,
  Matthieu Geist, and Olivier Bachem.
\newblock On-policy distillation of language models: Learning from
  self-generated mistakes.
\newblock \emph{arXiv preprint arXiv:2306.13649}, 2023.

\bibitem[Lu and Lab(2025)]{lu2025onpolicydistillation}
Kevin Lu and Thinking~Machines Lab.
\newblock On-policy distillation.
\newblock 2025.

\bibitem[Song and Zheng(2026)]{opdsurvey}
Mingyang Song and Mao Zheng.
\newblock A survey of on-policy distillation for large language models.
\newblock \emph{arXiv preprint arXiv:2604.00626}, 2026.

\bibitem[Fu et~al.(2026)Fu, Huang, Jiang, Zhu, and Zhao]{fu2026revisiting}
Yuqian Fu, Haohuan Huang, Kaiwen Jiang, Yuanheng Zhu, and Dongbin Zhao.
\newblock Revisiting on-policy distillation: Empirical failure modes and simple
  fixes.
\newblock \emph{arXiv preprint arXiv:2603.25562}, 2026.

\bibitem[Jin et~al.(2026)Jin, Min, Yang, Kadhe, Zhou, Wei, Baracaldo, and
  Lee]{jin2026entropy}
Woogyeol Jin, Taywon Min, Yongjin Yang, Swanand~Ravindra Kadhe, Yi~Zhou, Dennis
  Wei, Nathalie Baracaldo, and Kimin Lee.
\newblock Entropy-aware on-policy distillation of language models.
\newblock \emph{arXiv preprint arXiv:2603.07079}, 2026.

\bibitem[Jang et~al.(2026)Jang, Yeom, Yeo, Lim, and Kim]{jang2026stable}
Ijun Jang, Jewon Yeom, Juan Yeo, Hyunggu Lim, and Taesup Kim.
\newblock Stable on-policy distillation through adaptive target reformulation.
\newblock \emph{arXiv preprint arXiv:2601.07155}, 2026.

\bibitem[Li et~al.(2026{\natexlab{b}})Li, Yang, Fang, Song, Zheng, Guo, Zhang,
  Wang, and Chua]{li2026unifying}
Gengsheng Li, Tianyu Yang, Junfeng Fang, Mingyang Song, Mao Zheng, Haiyun Guo,
  Dan Zhang, Jinqiao Wang, and Tat-Seng Chua.
\newblock Unifying group-relative and self-distillation policy optimization via
  sample routing.
\newblock \emph{arXiv preprint arXiv:2604.02288}, 2026{\natexlab{b}}.

\bibitem[Ye et~al.(2026{\natexlab{a}})Ye, Dong, Wu, Huang, and
  Wei]{ye2026policy}
Tianzhu Ye, Li~Dong, Xun Wu, Shaohan Huang, and Furu Wei.
\newblock On-policy context distillation for language models.
\newblock \emph{arXiv preprint arXiv:2602.12275}, 2026{\natexlab{a}}.

\bibitem[Ye et~al.(2026{\natexlab{b}})Ye, Dong, Dong, Wu, Huang, and
  Wei]{ye2026online}
Tianzhu Ye, Li~Dong, Qingxiu Dong, Xun Wu, Shaohan Huang, and Furu Wei.
\newblock Online experiential learning for language models.
\newblock \emph{arXiv preprint arXiv:2603.16856}, 2026{\natexlab{b}}.

\bibitem[Zhao et~al.(2026{\natexlab{a}})Zhao, Xie, Liu, Huang, Pang, Chen, and
  Grover]{zhao2026self}
Siyan Zhao, Zhihui Xie, Mengchen Liu, Jing Huang, Guan Pang, Feiyu Chen, and
  Aditya Grover.
\newblock Self-distilled reasoner: On-policy self-distillation for large
  language models.
\newblock \emph{arXiv preprint arXiv:2601.18734}, 2026{\natexlab{a}}.

\bibitem[Zhao et~al.(2026{\natexlab{b}})Zhao, Xie, Wang, Li, Xie, and
  Sun]{zhao2026selfdistillationmultitokenprediction}
Guoliang Zhao, Ruobing Xie, An~Wang, Shuaipeng Li, Huaibing Xie, and Xingwu
  Sun.
\newblock Self-distillation for multi-token prediction, 2026.
\newblock \emph{arXiv preprint arXiv:2603.23911}, 2026{\natexlab{b}}.

\bibitem[H{\"u}botter et~al.(2026)H{\"u}botter, L{\"u}beck, Behric, Baumann,
  Bagatella, Marta, Hakimi, Shenfeld, Buening, Guestrin, and
  Krause]{rlselfdistill}
Jonas H{\"u}botter, Frederike L{\"u}beck, Lejs Behric, Anton Baumann, Marco
  Bagatella, Daniel Marta, Ido Hakimi, Idan Shenfeld, Thomas~Kleine Buening,
  Carlos Guestrin, and Andreas Krause.
\newblock {Reinforcement Learning via Self-Distillation}.
\newblock \emph{arXiv preprint arXiv:2601.20802}, 2026.
\newblock URL \url{https://arxiv.org/abs/2601.20802}.

\bibitem[Shenfeld et~al.(2026)Shenfeld, Damani, H\"ubotter, and
  Agrawal]{shenfeld2026self}
Idan Shenfeld, Mehul Damani, Jonas H\"ubotter, and Pulkit Agrawal.
\newblock Self-distillation enables continual learning.
\newblock \emph{arXiv preprint arXiv:2601.19897}, 2026.

\bibitem[Xu et~al.(2026)Xu, Sang, Zhou, He, and Wang]{paced}
Yuanda Xu, Hejian Sang, Zhengze Zhou, Ran He, and Zhipeng Wang.
\newblock {PACED: Distillation and On-Policy Self-Distillation at the Frontier
  of Student Competence}.
\newblock \emph{arXiv preprint arXiv:2603.11178}, 2026.
\newblock URL \url{https://arxiv.org/abs/2603.11178}.

\bibitem[Ko et~al.(2026)Ko, Abdali, Kim, Chen, and Cameron]{ko2026scaling}
Jongwoo Ko, Sara Abdali, Young~Jin Kim, Tianyi Chen, and Pashmina Cameron.
\newblock Scaling reasoning efficiently via relaxed on-policy distillation.
\newblock \emph{arXiv preprint arXiv:2603.11137}, 2026.

\bibitem[Kim et~al.(2026)Kim, Luo, Kim, Lee, Kim, Jeon, Li, and
  Yang]{kim2026does}
Jeonghye Kim, Xufang Luo, Minbeom Kim, Sangmook Lee, Dohyung Kim, Jiwon Jeon,
  Dongsheng Li, and Yuqing Yang.
\newblock Why does self-distillation (sometimes) degrade the reasoning
  capability of llms?
\newblock \emph{arXiv preprint arXiv:2603.24472}, 2026.

\bibitem[Ye et~al.(2025)Ye, Dong, Chi, Wu, Huang, and
  Wei]{zhang2025blackboxopd}
Tianzhu Ye, Li~Dong, Zewen Chi, Xun Wu, Shaohan Huang, and Furu Wei.
\newblock {Black-Box On-Policy Distillation of Large Language Models}.
\newblock \emph{arXiv preprint arXiv:2511.10643}, 2025.
\newblock URL \url{https://arxiv.org/abs/2511.10643}.

\bibitem[Xiao et~al.(2026)Xiao, Xia, Yang, Gao, Shen, Zhang, He, Lou, Luo,
  Wang, et~al.]{xiao2026mimo}
Bangjun Xiao, Bingquan Xia, Bo~Yang, Bofei Gao, Bowen Shen, Chen Zhang,
  Chenhong He, Chiheng Lou, Fuli Luo, Gang Wang, et~al.
\newblock Mimo-v2-flash technical report.
\newblock \emph{arXiv preprint arXiv:2601.02780}, 2026.

\bibitem[Xu et~al.(2025{\natexlab{b}})Xu, Pang, Zhu, Gu, Wei, Deng, Pan, Shen,
  and Cheng]{rlkd}
Shicheng Xu, Liang Pang, Yunchang Zhu, Jia Gu, Zihao Wei, Jingcheng Deng,
  Feiyang Pan, Huawei Shen, and Xueqi Cheng.
\newblock {RLKD: Distilling LLMs' Reasoning via Reinforcement Learning}.
\newblock \emph{arXiv preprint arXiv:2505.16142}, 2025{\natexlab{b}}.
\newblock URL \url{https://arxiv.org/abs/2505.16142}.

\bibitem[Xu et~al.(2025{\natexlab{c}})Xu, Zhu, Deng, Li, Hou, Wang, Shang, Xu,
  and Mi]{kdrl}
Hongling Xu, Qi~Zhu, Heyuan Deng, Jinpeng Li, Lu~Hou, Yasheng Wang, Lifeng
  Shang, Ruifeng Xu, and Fei Mi.
\newblock {KDRL: Post-Training Reasoning LLMs via Unified Knowledge
  Distillation and Reinforcement Learning}.
\newblock \emph{arXiv preprint arXiv:2506.02208}, 2025{\natexlab{c}}.
\newblock URL \url{https://arxiv.org/abs/2506.02208}.

\bibitem[Zhang et~al.(2026{\natexlab{b}})Zhang, Jiang, Shen, Zhang, Ram, Yang,
  Tu, Xia, and Soatto]{rakd}
Zhaoyang Zhang, Shuli Jiang, Yantao Shen, Yuting Zhang, Dhananjay Ram, Shuo
  Yang, Zhuowen Tu, Wei Xia, and Stefano Soatto.
\newblock {Reinforcement-aware Knowledge Distillation for LLM Reasoning}.
\newblock \emph{arXiv preprint arXiv:2602.22495}, 2026{\natexlab{b}}.
\newblock URL \url{https://arxiv.org/abs/2602.22495}.

\bibitem[Zhang et~al.(2025)Zhang, Zhang, Zhang, Hu, Chen, and Xu]{aligndistil}
Songming Zhang, Xue Zhang, Tong Zhang, Bojie Hu, Yufeng Chen, and Jinan Xu.
\newblock {AlignDistil: Token-Level Language Model Alignment as Adaptive Policy
  Distillation}.
\newblock In \emph{Proceedings of ACL}, 2025.
\newblock URL \url{https://arxiv.org/abs/2503.02832}.

\bibitem[Yang et~al.(2026{\natexlab{a}})Yang, Qin, Si, Chen, Gu, Yao, Lin,
  Wang, Wang, and Duan]{yang2026self}
Chenxu Yang, Chuanyu Qin, Qingyi Si, Minghui Chen, Naibin Gu, Dingyu Yao, Zheng
  Lin, Weiping Wang, Jiaqi Wang, and Nan Duan.
\newblock Self-distilled rlvr.
\newblock \emph{arXiv preprint arXiv:2604.03128}, 2026{\natexlab{a}}.

\bibitem[Li et~al.(2025{\natexlab{b}})Li, Yue, Xu, Jiang, Niu, Lin,
  Ramasubramanian, and Poovendran]{li2025small}
Yuetai Li, Xiang Yue, Zhangchen Xu, Fengqing Jiang, Luyao Niu, Bill~Yuchen Lin,
  Bhaskar Ramasubramanian, and Radha Poovendran.
\newblock Small models struggle to learn from strong reasoners.
\newblock 2025{\natexlab{b}}.

\bibitem[Yang et~al.(2026{\natexlab{b}})Yang, Liu, Xie, Yang, Yang, and
  Lin]{gopd}
Wenkai Yang, Weijie Liu, Ruobing Xie, Kai Yang, Saiyong Yang, and Yankai Lin.
\newblock Learning beyond teacher: Generalized on-policy distillation with
  reward extrapolation.
\newblock \emph{arXiv preprint arXiv:2602.12125}, 2026{\natexlab{b}}.

\bibitem[Burns et~al.(2023)Burns, Izmailov, Kirchner, Baker, Gao,
  Aschenbrenner, Chen, Ecoffet, Joglekar, Leike, Sutskever, and
  Wu]{burns2023weakstrong}
Collin Burns, Pavel Izmailov, Jan~Hendrik Kirchner, Bowen Baker, Leo Gao,
  Leopold Aschenbrenner, Yining Chen, Adrien Ecoffet, Manas Joglekar, Jan
  Leike, Ilya Sutskever, and Jeff Wu.
\newblock Weak-to-strong generalization: Eliciting strong capabilities with
  weak supervision.
\newblock \emph{arXiv preprint arXiv:2312.09390}, 2023.

\bibitem[Leike and Sutskever(2023)]{superalignment}
Jan Leike and Ilya Sutskever.
\newblock {Introducing Superalignment}.
\newblock \emph{OpenAI Blog}, 2023.

\bibitem[Grandvalet and Bengio(2004)]{grandvalet2004semi}
Yves Grandvalet and Yoshua Bengio.
\newblock {Semi-supervised learning by entropy minimization}.
\newblock \emph{Advances in neural information processing systems}, 17, 2004.

\bibitem[Dai and Le(2015)]{dai2015semi}
Andrew~M Dai and Quoc~V Le.
\newblock {Semi-supervised sequence learning}.
\newblock \emph{Advances in neural information processing systems}, 28, 2015.

\bibitem[Laine and Aila(2016)]{laine2016temporal}
Samuli Laine and Timo Aila.
\newblock {Temporal ensembling for semi-supervised learning}.
\newblock \emph{arXiv preprint arXiv:1610.02242}, 2016.

\bibitem[Athiwaratkun et~al.(2018)Athiwaratkun, Finzi, Izmailov, and
  Wilson]{athiwaratkun2018there}
Ben Athiwaratkun, Marc Finzi, Pavel Izmailov, and Andrew~Gordon Wilson.
\newblock {There are many consistent explanations of unlabeled data: Why you
  should average}.
\newblock \emph{arXiv preprint arXiv:1806.05594}, 2018.

\bibitem[Han et~al.(2018)Han, Yao, Yu, Niu, Xu, Hu, Tsang, and
  Sugiyama]{han2018co}
Bo~Han, Quanming Yao, Xingrui Yu, Gang Niu, Miao Xu, Weihua Hu, Ivor Tsang, and
  Masashi Sugiyama.
\newblock {Co-teaching: Robust training of deep neural networks with extremely
  noisy labels}.
\newblock \emph{Advances in neural information processing systems}, 31, 2018.

\bibitem[Berthelot et~al.(2019)Berthelot, Carlini, Goodfellow, Papernot,
  Oliver, and Raffel]{berthelot2019mixmatch}
David Berthelot, Nicholas Carlini, Ian Goodfellow, Nicolas Papernot, Avital
  Oliver, and Colin~A Raffel.
\newblock {Mixmatch: A holistic approach to semi-supervised learning}.
\newblock \emph{Advances in neural information processing systems}, 32, 2019.

\bibitem[Li et~al.(2020)Li, Socher, and Hoi]{li2020dividemix}
Junnan Li, Richard Socher, and Steven~CH Hoi.
\newblock {Dividemix: Learning with noisy labels as semi-supervised learning}.
\newblock \emph{arXiv preprint arXiv:2002.07394}, 2020.

\bibitem[Chen et~al.(2020{\natexlab{a}})Chen, Kornblith, Swersky, Norouzi, and
  Hinton]{chen2020big}
Ting Chen, Simon Kornblith, Kevin Swersky, Mohammad Norouzi, and Geoffrey
  Hinton.
\newblock {Big self-supervised models are strong semi-supervised learners}.
\newblock \emph{Advances in neural information processing systems},
  33:\penalty0 22243--22255, 2020{\natexlab{a}}.

\bibitem[Chen et~al.(2020{\natexlab{b}})Chen, Wei, Kumar, and Ma]{chen2020self}
Yining Chen, Colin Wei, Ananya Kumar, and Tengyu Ma.
\newblock {Self-training avoids using spurious features under domain shift}.
\newblock \emph{Advances in Neural Information Processing Systems},
  33:\penalty0 21061--21071, 2020{\natexlab{b}}.

\bibitem[Chen et~al.(2022)Chen, Jiang, Wang, Wan, Wang, and
  Long]{chen2022debiased}
Baixu Chen, Junguang Jiang, Ximei Wang, Pengfei Wan, Jianmin Wang, and
  Mingsheng Long.
\newblock {Debiased self-training for semi-supervised learning}.
\newblock \emph{Advances in Neural Information Processing Systems},
  35:\penalty0 32424--32437, 2022.

\bibitem[Christiano et~al.(2018)Christiano, Shlegeris, and
  Amodei]{christiano2018supervising}
Paul Christiano, Buck Shlegeris, and Dario Amodei.
\newblock {Supervising strong learners by amplifying weak experts}.
\newblock \emph{arXiv preprint arXiv:1810.08575}, 2018.

\bibitem[Leike et~al.(2018)Leike, Krueger, Everitt, Martic, Maini, and
  Legg]{leike2018scalable}
Jan Leike, David Krueger, Tom Everitt, Miljan Martic, Vishal Maini, and Shane
  Legg.
\newblock Scalable agent alignment via reward modeling: a research direction.
\newblock \emph{arXiv preprint arXiv:1811.07871}, 2018.

\bibitem[Irving et~al.(2018)Irving, Christiano, and Amodei]{irving2018debate}
Geoffrey Irving, Paul Christiano, and Dario Amodei.
\newblock {AI} safety via debate.
\newblock \emph{arXiv preprint arXiv:1805.00899}, 2018.

\bibitem[Bowman et~al.(2022)Bowman, Hyun, Perez, Chen, Pettit, Heiner,
  Lukosuite, Askell, Jones, Chen, et~al.]{bowman2022scalable}
Samuel~R Bowman, Jeeyoon Hyun, Ethan Perez, Edwin Chen, Craig Pettit, Scott
  Heiner, Kamile Lukosuite, Amanda Askell, Andy Jones, Anna Chen, et~al.
\newblock Measuring progress on scalable oversight for large language models.
\newblock \emph{arXiv preprint arXiv:2211.03540}, 2022.

\bibitem[Bowman(2022)]{bowman2022alignment}
Sam Bowman.
\newblock {Artificial Sandwiching: When can we test scalable alignment
  protocols without humans?}
\newblock \emph{AI Alignment Forum}, 2022.

\bibitem[Bai et~al.(2022)Bai, Kadavath, Kundu, Askell, Kernion, Jones, Chen,
  Goldie, Mirhoseini, McKinnon, et~al.]{bai2022constitutional}
Yuntao Bai, Saurav Kadavath, Sandipan Kundu, Amanda Askell, Jackson Kernion,
  Andy Jones, Anna Chen, Anna Goldie, Azalia Mirhoseini, Cameron McKinnon,
  et~al.
\newblock Constitutional {AI}: Harmlessness from {AI} feedback.
\newblock \emph{arXiv preprint arXiv:2212.08073}, 2022.

\bibitem[Christiano et~al.(2022)Christiano, Cotra, and
  Xu]{christiano2022eliciting}
Paul Christiano, Ajeya Cotra, and Mark Xu.
\newblock {Eliciting latent knowledge}.
\newblock Technical report, Alignment Research Center (ARC), 2022.

\bibitem[Burns et~al.(2022)Burns, Ye, Klein, and
  Steinhardt]{burns2022discovering}
Collin Burns, Haotian Ye, Dan Klein, and Jacob Steinhardt.
\newblock {Discovering latent knowledge in language models without
  supervision}.
\newblock \emph{arXiv preprint arXiv:2212.03827}, 2022.

\bibitem[Schwarzschild et~al.(2021{\natexlab{a}})Schwarzschild, Borgnia, Gupta,
  Huang, Vishkin, Goldblum, and Goldstein]{schwarzschild2021can}
Avi Schwarzschild, Eitan Borgnia, Arjun Gupta, Furong Huang, Uzi Vishkin, Micah
  Goldblum, and Tom Goldstein.
\newblock {Can you learn an algorithm? generalizing from easy to hard problems
  with recurrent networks}.
\newblock \emph{Advances in Neural Information Processing Systems},
  34:\penalty0 6695--6706, 2021{\natexlab{a}}.

\bibitem[Schwarzschild et~al.(2021{\natexlab{b}})Schwarzschild, Borgnia, Gupta,
  Bansal, Emam, Huang, Goldblum, and Goldstein]{schwarzschild2021datasets}
Avi Schwarzschild, Eitan Borgnia, Arjun Gupta, Arpit Bansal, Zeyad Emam, Furong
  Huang, Micah Goldblum, and Tom Goldstein.
\newblock {Datasets for studying generalization from easy to hard examples}.
\newblock \emph{arXiv preprint arXiv:2108.06011}, 2021{\natexlab{b}}.

\bibitem[Cao et~al.(2024)Cao, Lu, Lu, Chen, Ren, Xiang, Liu, Lu, He, Han, Sun,
  Lin, et~al.]{survey_autoalign}
Boxi Cao, Keming Lu, Xinyu Lu, Jiawei Chen, Mengjie Ren, Hao Xiang, Peilin Liu,
  Yaojie Lu, Ben He, Xianpei Han, Le~Sun, Hongyu Lin, et~al.
\newblock Towards scalable automated alignment of llms: A survey.
\newblock \emph{arXiv preprint arXiv:2406.01252}, 2024.

\bibitem[Yao et~al.(2025{\natexlab{a}})Yao, Yang, Wang, Lin, and
  Liu]{bansal2025revisiting_w2s}
Wei Yao, Wenkai Yang, Ziqiao Wang, Yankai Lin, and Yong Liu.
\newblock Revisiting weak-to-strong generalization in theory and practice:
  Reverse kl vs. forward kl.
\newblock \emph{arXiv preprint arXiv:2502.11107}, 2025{\natexlab{a}}.

\bibitem[Yao et~al.(2025{\natexlab{b}})Yao, Xu, Tang, Yang, Di, Wang, and
  Liu]{w2s_fdivergence}
Wei Yao, Gengze Xu, Huayi Tang, Wenkai Yang, Donglin Di, Ziqiao Wang, and Yong
  Liu.
\newblock On weak-to-strong generalization and f-divergence.
\newblock \emph{arXiv preprint arXiv:2506.03109}, 2025{\natexlab{b}}.

\bibitem[Sang et~al.(2024)Sang, Wang, Zhang, Zhu, Kong, Ye, Wei, and
  Xiao]{guo2024improving_w2s}
Jitao Sang, Yuhang Wang, Jing Zhang, Yanxu Zhu, Chao Kong, Junhong Ye, Shuyu
  Wei, and Jinlin Xiao.
\newblock Improving weak-to-strong generalization with scalable oversight and
  ensemble learning.
\newblock \emph{arXiv preprint arXiv:2402.00667}, 2024.

\bibitem[Yuan et~al.(2025)Yuan, Xiao, Tao, Wang, Gao, Ding, and
  Xu]{w2s_reasoning}
Yige Yuan, Teng Xiao, Shuchang Tao, Xue Wang, Jinyang Gao, Bolin Ding, and
  Bingbing Xu.
\newblock Incentivizing strong reasoning from weak supervision.
\newblock \emph{arXiv preprint arXiv:2505.20072}, 2025.

\bibitem[Yuan et~al.(2024)Yuan, Pang, Cho, Sukhbaatar, Xu, and
  Weston]{yuan2024self}
Weizhe Yuan, Richard~Yuanzhe Pang, Kyunghyun Cho, Sainbayar Sukhbaatar, Jing
  Xu, and Jason Weston.
\newblock Self-rewarding language models.
\newblock \emph{arXiv preprint arXiv:2401.10020}, 2024.

\bibitem[Zhu et~al.(2025)Zhu, He, Wang, Liu, and Wang]{wspo}
Wenhong Zhu, Zhiwei He, Xiaofeng Wang, Pengfei Liu, and Rui Wang.
\newblock Weak-to-strong preference optimization: Stealing reward from weak
  aligned model.
\newblock \emph{arXiv preprint arXiv:2410.18640}, 2025.

\bibitem[Azar et~al.(2024)Azar, Rowland, Piot, Guo, Calandriello, Valko, and
  Munos]{azar2024ipo}
Mohammad~Gheshlaghi Azar, Mark Rowland, Bilal Piot, Daniel Guo, Daniele
  Calandriello, Michal Valko, and R{\'e}mi Munos.
\newblock A general theoretical paradigm to understand learning from human
  preferences.
\newblock \emph{AISTATS}, 2024.

\bibitem[Ethayarajh et~al.(2024)Ethayarajh, Xu, Muennighoff, Jurafsky, and
  Kiela]{ethayarajh2024kto}
Kawin Ethayarajh, Winnie Xu, Niklas Muennighoff, Dan Jurafsky, and Douwe Kiela.
\newblock {KTO}: Model alignment as prospect theoretic optimization.
\newblock \emph{arXiv preprint arXiv:2402.01306}, 2024.

\bibitem[Meng et~al.(2024)Meng, Xia, and Chen]{meng2024simpo}
Yu~Meng, Mengzhou Xia, and Danqi Chen.
\newblock {SimPO}: Simple preference optimization with a reference-free reward.
\newblock \emph{NeurIPS}, 2024.

\bibitem[Lightman et~al.(2023)Lightman, Kosaraju, Burda, Edwards, Baker, Lee,
  Leike, Schulman, Sutskever, and Cobbe]{lightman2023let}
Hunter Lightman, Vineet Kosaraju, Yura Burda, Harri Edwards, Bowen Baker, Teddy
  Lee, Jan Leike, John Schulman, Ilya Sutskever, and Karl Cobbe.
\newblock Let's verify step by step.
\newblock \emph{arXiv preprint arXiv:2305.20050}, 2023.

\bibitem[Wang et~al.(2023)Wang, Li, Shao, Xu, Dai, Li, Chen, Wu, and
  Sui]{mathshepherd}
Peiyi Wang, Lei Li, Zhihong Shao, RX~Xu, Damai Dai, Yifei Li, Deli Chen, Y~Wu,
  and Zhifang Sui.
\newblock Math-shepherd: A label-free step-by-step verifier for {LLMs} in
  mathematical reasoning.
\newblock \emph{arXiv preprint arXiv:2312.08935}, 2023.

\bibitem[Cui et~al.(2025)Cui, Yuan, Wang, Wang, Li, He, Fan, Yu, Xu, Chen,
  et~al.]{cui2025process}
Ganqu Cui, Lifan Yuan, Zefan Wang, Hanbin Wang, Wendi Li, Bingxiang He, Yuchen
  Fan, Tianyu Yu, Qixin Xu, Weize Chen, et~al.
\newblock Process reinforcement through implicit rewards.
\newblock \emph{arXiv preprint arXiv:2502.01456}, 2025.

\bibitem[Kazemnejad et~al.(2024)Kazemnejad, Aghajohari, Portelance, Sordoni,
  Reddy, Courville, and Le~Roux]{kazemnejad2024vineppo}
Amirhossein Kazemnejad, Milad Aghajohari, Eva Portelance, Alessandro Sordoni,
  Siva Reddy, Aaron Courville, and Nicolas Le~Roux.
\newblock {VinePPO}: Unlocking rl potential for llm reasoning through refined
  credit assignment.
\newblock \emph{arXiv preprint arXiv:2410.01679}, 2024.

\bibitem[Zhang et~al.(2026{\natexlab{c}})]{creditassignsurvey}
Chenchen Zhang et~al.
\newblock From reasoning to agentic: Credit assignment in reinforcement
  learning for large language models.
\newblock \emph{arXiv preprint arXiv:2604.09459}, 2026{\natexlab{c}}.

\bibitem[Gao et~al.(2023)Gao, Schulman, and Hilton]{gao2023overopt}
Leo Gao, John Schulman, and Jacob Hilton.
\newblock Scaling laws for reward model overoptimization.
\newblock \emph{ICML}, 2023.

\bibitem[Ilharco et~al.(2023)Ilharco, Ribeiro, Wortsman, Schmidt, Hajishirzi,
  and Farhadi]{ilharco2023task}
Gabriel Ilharco, Marco~Tulio Ribeiro, Mitchell Wortsman, Ludwig Schmidt,
  Hannaneh Hajishirzi, and Ali Farhadi.
\newblock Editing models with task arithmetic.
\newblock In \emph{ICLR}, 2023.

\bibitem[Wortsman et~al.(2022)Wortsman, Ilharco, Gadre, Roelofs, Gontijo-Lopes,
  Morcos, Namkoong, Farhadi, Carmon, Kornblith, et~al.]{wortsman2022soups}
Mitchell Wortsman, Gabriel Ilharco, Samir~Yitzhak Gadre, Rebecca Roelofs,
  Raphael Gontijo-Lopes, Ari~S Morcos, Hongseok Namkoong, Ali Farhadi, Yair
  Carmon, Simon Kornblith, et~al.
\newblock Model soups: Averaging weights of multiple fine-tuned models improves
  accuracy without increasing inference time.
\newblock In \emph{ICML}, 2022.

\bibitem[Liu et~al.(2024)Liu, Han, Wang, Tsvetkov, Choi, and
  Smith]{proxytuning}
Alisa Liu, Xiaochuang Han, Yizhong Wang, Yulia Tsvetkov, Yejin Choi, and
  Noah~A. Smith.
\newblock Tuning language models by proxy.
\newblock \emph{arXiv preprint arXiv:2401.08565}, 2024.

\end{thebibliography}
\bibliographystyle{unsrtnat}

\newpage
\appendix
\renewcommand{\thesection}{\Alph{section}}
\titleformat{\section}
  {\sectionfont\sffamily\bfseries\gensicolor}
  {\thesection.}
  {0.5em}
  {}

\section{Experimental Details}



\paragraph{Direct-OPD data and prompt.}
For the final \method{} runs, we use the math subset of Skywork-OR1-RL-Data,
released with Skywork-OR1~\citep{he2025skywork}, and apply the following
DAPO-style math prompt template.

\begin{mdframed}[
  skipabove=0.45em,
  skipbelow=0.45em,
  innerleftmargin=8pt,
  innerrightmargin=8pt,
  innertopmargin=6pt,
  innerbottommargin=6pt
]
\footnotesize\ttfamily
Solve the following math problem step by step.\\
The last line of your response should be of the form\\
Answer: \$Answer (without quotes) where \$Answer is the answer to the problem.\\
\\
\{Question\}\\
\\
Remember to put your answer on its own line after "Answer:".
\end{mdframed}
We use this template for \method{} training rollouts and evaluation prompts.
This template differs from the boxed-answer prompt used during teacher RL; in
our runs, the DAPO-style prompt gives slightly better transfer. Since prompt
template divergence is not the focus of this paper, we use this setting
throughout. We also observe similar transfer trends when replacing the Skywork
training dataset with DAPO-Math-17K while keeping the same prompt format,
suggesting that the result is not specific to a single training dataset.

\paragraph{Evaluation.}
Table~\ref{tab:eval-settings} summarizes the evaluation protocol used for all
reported AIME results unless otherwise specified.

\begin{table}[H]
\centering
\small
\setlength{\tabcolsep}{6pt}
\renewcommand{\arraystretch}{1.08}
\begin{tabular}{lc}
\toprule
Evaluation setting & Value \\
\midrule
Benchmarks & AIME 2024, AIME 2025 \\
Samples per problem & 32 \\
Sampling temperature & 0.7 \\
Top-$p$ & 0.95 \\
Maximum generation length & 31,744 \\
\bottomrule
\end{tabular}
\caption{Evaluation protocol for AIME validation.}
\label{tab:eval-settings}
\end{table}

Table~\ref{tab:dopd-training-hyperparams} lists the default \method{} training
hyperparameters used with the data and prompt format above.

\begin{table}[H]
\centering
\small
\setlength{\tabcolsep}{6pt}
\renewcommand{\arraystretch}{1.08}
\begin{tabular}{lc}
\toprule
Hyperparameter & Value \\
\midrule
Training framework & \texttt{verl} \\
Global batch size & 64 \\
Mini batch size & 64 \\
Rollout $n$ & 4 \\
Maximum prompt length & 1,024 \\
Maximum response length & 2,048 \\
Sampling temperature & 1.0 \\
Top-$p$ & 1.0 \\
Learning rate & $1\times10^{-6}$ \\
Training steps & 300 \\
KL coefficient & $[0.8, 2]$ for different student--teacher pairs if not using adaptive KL \\
Student top-$k$ support & 16 \\
Top-$k$ strategy & Student top-$k$ \\
\bottomrule
\end{tabular}
\caption{Training hyperparameters for \method{} unless otherwise specified.}
\label{tab:dopd-training-hyperparams}
\end{table}

Table~\ref{tab:rl-training-settings} reports the GRPO settings used to train
RL teachers and direct-RL baselines.

\begin{table}[H]
\centering
\small
\setlength{\tabcolsep}{4.5pt}
\renewcommand{\arraystretch}{1.06}
\begin{minipage}[t]{0.48\linewidth}
\centering
\textbf{R1-Distill-1.5B RL}\\[3pt]
\begin{tabular}{lc}
\toprule
Hyperparameter & Value \\
\midrule
Algorithm & GRPO \\
Train batch size & 512 \\
PPO mini batch size & 128 \\
Rollout $n$ & 8 \\
Maximum prompt length & 2,048 \\
Maximum response length & 16,384 \\
Temperature & 1.0 \\
Learning rate & $1\times10^{-6}$ \\
KL coefficient & 0.0 \\
Clip high & 0.28 \\
Clip low & 0.2 \\
\bottomrule
\end{tabular}
\end{minipage}\hfill
\begin{minipage}[t]{0.48\linewidth}
\centering
\textbf{R1-Distill-7B RL}\\[3pt]
\begin{tabular}{lc}
\toprule
Hyperparameter & Value \\
\midrule
Algorithm & GRPO \\
Train batch size & 512 \\
PPO mini batch size & 128 \\
Rollout $n$ & 8 \\
Maximum prompt length & 2,048 \\
Maximum response length & 16,384 \\
Temperature & 1.0 \\
Learning rate & $1\times10^{-6}$ \\
KL coefficient & 0.0 \\
Clip high & 0.28 \\
Clip low & 0.2 \\
\bottomrule
\end{tabular}
\end{minipage}

\vspace{0.8em}

\begin{minipage}[t]{0.48\linewidth}
\centering
\textbf{Qwen3-1.7B-nonthinking RL}\\[3pt]
\begin{tabular}{lc}
\toprule
Hyperparameter & Value \\
\midrule
Algorithm & GRPO \\
Train batch size & 128 \\
PPO mini batch size & 128 \\
Rollout $n$ & 8 \\
Maximum prompt length & 2,048 \\
Maximum response length & 16,384 \\
Temperature & 1.0 \\
Learning rate & $1\times10^{-6}$ \\
KL coefficient & 0.0 \\
\bottomrule
\end{tabular}
\end{minipage}\hfill
\begin{minipage}[t]{0.48\linewidth}
\centering
\textbf{Qwen3-4B-nonthinking RL}\\[3pt]
\begin{tabular}{lc}
\toprule
Hyperparameter & Value \\
\midrule
Algorithm & GRPO \\
Train batch size & 128 \\
PPO mini batch size & 128 \\
Rollout $n$ & 8 \\
Maximum prompt length & 2,048 \\
Maximum response length & 16,384 \\
Temperature & 1.0 \\
Learning rate & $1\times10^{-6}$ \\
KL coefficient & 0.0 \\
\bottomrule
\end{tabular}
\end{minipage}
\caption{RL training settings for teacher construction and direct-RL
baselines. We follow the math GRPO setting of \citet{gopd}, using train batch
size 512 for R1-Distill runs, train batch size 128 for Qwen3-nonthinking runs,
and PPO mini batch size 128 throughout.}
\label{tab:rl-training-settings}
\end{table}

%
%

\section{Additional Entropy Diagnostics}

\begin{figure}[H]
\centering
\includegraphics[width=\linewidth]{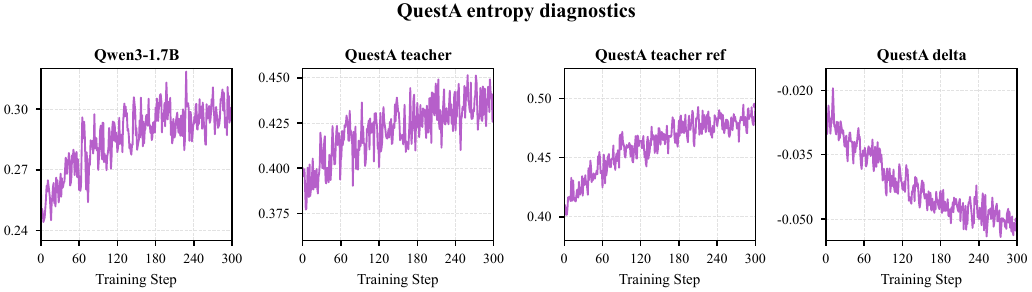}
\caption{Entropy diagnostics for QuestA-Nemotron into Qwen3-1.7B.
The panels show student entropy, post-RL teacher entropy, teacher-reference
entropy, and teacher entropy minus reference entropy. Together with
Figure~\ref{fig:qwen3-entropy}, this shows that the non-collapse pattern is not
specific to the JustRL teacher pair.}
\label{fig:questa-qwen3-entropy-appendix}
\end{figure}

\section{Additional Results}

\begin{figure}[H]
\centering
\includegraphics[width=0.92\linewidth]{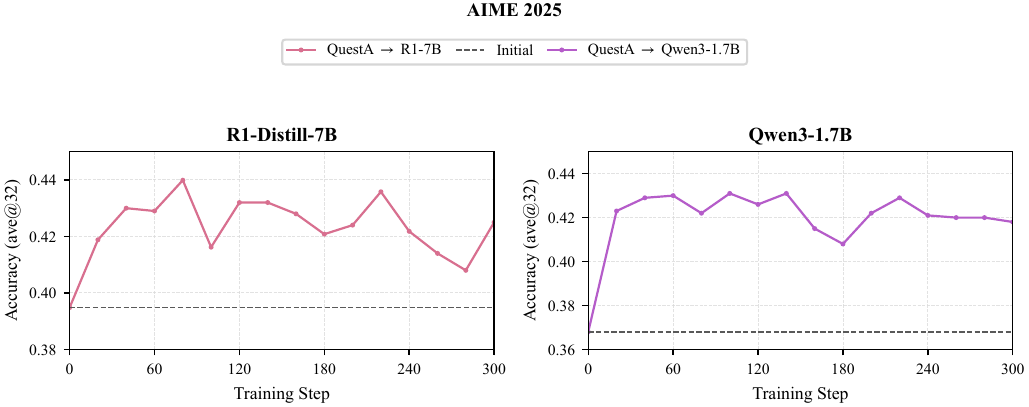}
\caption{QuestA transfer curves on AIME 2025 for the cross-pattern transfer
setting. The main text reports the corresponding AIME 2024 curves.}
\label{fig:questa-transfer-aime25-appendix}
\end{figure}

\end{document}